\documentclass[10pt,twocolumn,letterpaper]{article}

\usepackage{cvpr}
\usepackage{times}
\usepackage{epsfig}
\usepackage{graphicx}
\usepackage{amsmath}
\usepackage{amssymb}
\usepackage{mathtools}
\usepackage{multirow}
\usepackage{wrapfig}
\usepackage{subcaption}

\usepackage{ dsfont }
\usepackage{cuted}
\usepackage{capt-of}
\usepackage{verbatim} 
\usepackage[title]{appendix}

\usepackage[pagebackref=true,breaklinks=true,letterpaper=true,colorlinks,bookmarks=false]{hyperref}

\cvprfinalcopy 

\def\cvprPaperID{821} 

\newcommand*{\affaddr}[1]{#1} 
\newcommand*{\affmark}[1][*]{\textsuperscript{#1}}
\graphicspath{{figures-arxiv/}}
\ifcvprfinal\fi

\begin{document}

\title{GANFIT: Generative Adversarial Network Fitting \\for High Fidelity 3D Face Reconstruction}

\author{
Baris Gecer\affmark[1,2], Stylianos Ploumpis\affmark[1,2], Irene Kotsia\affmark[3], and Stefanos Zafeiriou\affmark[1,2]\\
\affaddr{\affmark[1]Imperial College London}\\
\affaddr{\affmark[2]FaceSoft.io}\\
\affaddr{\affmark[3]University of Middlesex}\\
{\tt\small \{b.gecer, s.ploumpis, s.zafeiriou\}@imperial.ac.uk}
{\tt\small , drkotsia@gmail.com}}


\maketitle


\begin{strip}\centering
\vspace{-1cm}
\includegraphics[width=1\textwidth]{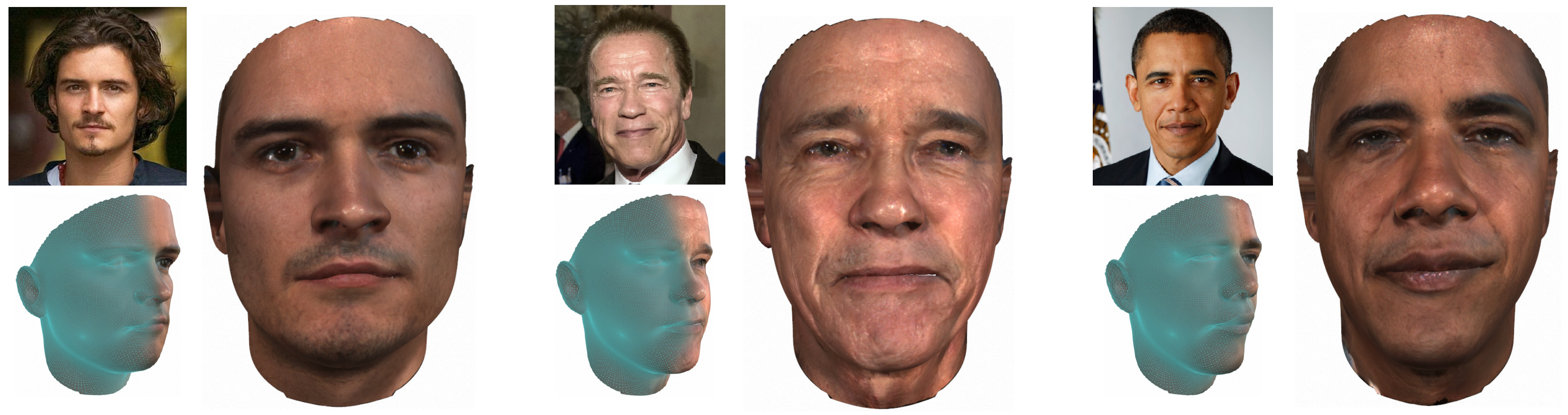}
\captionof{figure}{The proposed deep fitting approach can reconstruct high quality texture and geometry from a single image with precise identity recovery. The reconstructions in the figure and the rest of the paper are represented by a vector of size 700 floating points and rendered without any special effects. We would like to highlight that the depicted texture is reconstructed by our model and none of the features taken directly from the image. 
\label{fig:cover}}
\end{strip}

\begin{abstract}
In the past few years, a lot of work has been done towards reconstructing the 3D facial structure from single images by capitalizing on the power of Deep Convolutional Neural Networks (DCNNs). In the most recent works, differentiable renderers were employed in order to learn the relationship between the facial identity features and the parameters of a 3D morphable model for shape and texture. The texture features either correspond to components of a linear texture space or are learned by auto-encoders directly from in-the-wild images. In all cases, the quality of the facial texture reconstruction of the state-of-the-art methods is still not capable of modeling textures in  high fidelity. In this paper, we take a radically different approach and harness the power of Generative Adversarial Networks (GANs) and DCNNs in order to reconstruct the facial texture and shape from single images. That is, we utilize GANs to train a very powerful generator of facial texture in UV space. Then, we revisit the original 3D Morphable Models (3DMMs) fitting approaches making use of non-linear optimization to find the optimal latent parameters that best reconstruct the test image but under a new perspective. We optimize the parameters with the supervision of pretrained deep identity features through our end-to-end differentiable framework. We demonstrate excellent results in photorealistic and identity preserving 3D face reconstructions and achieve for the first time, to the best of our knowledge, facial texture reconstruction with high-frequency details.\footnote{Project page: \url{https://github.com/barisgecer/ganfit}} 

\end{abstract}

\section{Introduction}
Estimation of the 3D facial surface and other intrinsic components of the face from single images (e.g., albedo, etc.) is a very important problem at the intersection of computer vision and machine learning with countless applications (e.g., face recognition, face editing, virtual reality). It is now twenty years from the seminal work of Blanz and Vetter~\cite{blanz1999morphable} which showed that it is possible to reconstruct shape and albedo by solving a non-linear optimization problem that is constrained by linear statistical models of facial texture and shape. This statistical model of texture and shape is called a 3D Morphable Model (3DMM). Arguably the most popular publicly available 3DMM is the Basel model built from 200 people~\cite{bfm09}. Recently, large scale statistical models of face and head shape have been made publicly available~\cite{booth20163d,dai20173d}. 

For many years 3DMMs and its variants were the methods of choice for 3D face reconstruction ~\cite{romdhani2005estimating,zhu2016face,jiang20183d}. Furthermore, with appropriate statistical texture models on image features such as Scale Invariant Feature Transform (SIFT) and Histogram Of Gradients (HOG), 3DMM-based methodologies can still achieve state-of-the-art performance in 3D shape estimation on images captured under unconstrained conditions~\cite{booth20173d}. Nevertheless, those methods~\cite{booth20173d} can reconstruct only the shape and not the facial texture. Another line of research in \cite{yamaguchi2018high,saito2017photorealistic} decouples texture and shape reconstruction. A standard linear 3DMM fitting strategy~\cite{thies2016face2face} is used for face reconstruction followed by a number of steps for texture completion and refinement. In these papers~\cite{saito2017photorealistic,yamaguchi2018high}, the texture looks excellent when rendered under professional renderers (e.g., Arnold), nevertheless when the texture is overlaid on the images the quality  significantly drops \footnote{Please see the supplementary materials for a comparison with \cite{saito2017photorealistic,yamaguchi2018high}.}.

In the past two years, a lot of work has been conducted on how to harness Deep Convolutional Neural Networks (DCNNs) for 3D shape and texture reconstruction. The first such methods either trained regression DCNNs from image to the parameters of a 3DMM~\cite{tran2017regressing} or used a 3DMM to synthesize images~\cite{richardson20163d,guo2018cnn} and formulate an image-to-image translation problem using DCNNs to estimate the depth\footnote{The depth was afterwards refined by fitting a 3DMM and then changing the normals by using image features.}~\cite{sela2017unrestricted}. The more recent unsupervised DCNN-based methods are trained to regress 3DMM parameters from identity features by making use of differentiable image formation architectures~\cite{cole2017synthesizing} and differentiable renderers ~\cite{genova2018unsupervised,tewari2017mofa,richardson2017learning}. 

The most recent methods such as~\cite{tewari2017self,tran2018nonlinear,garrido2016reconstruction} use both the 3DMM model, as well as additional network structures (called correctives) in order to extend the shape and texture representation. Even though the paper~\cite{tewari2017self} shows that the reconstructed facial texture has indeed more details than a texture estimated from a 3DMM~\cite{tran2017regressing,tewari2017mofa}, it is still unable to capture high-frequency details in texture and subsequently many identity characteristics (please see the Fig.~\ref{fig:comparison}). Furthermore, because the method permits the reconstructions to be  outside the 3DMM space, it is susceptible to outliers (e.g., glasses etc.) which are baked in shape and texture. Although rendering networks (\ie trained by VAE~\cite{lombardi2018deep}) generates outstanding quality textures, each network is capable of storing up to few individuals whom should be placed in a controlled environment to collect ${\sim}20$ millions of images.

In this paper, we still propose to build upon the success of DCNNs but take a radically different approach for 3D shape and texture reconstruction from a single in-the-wild image. That is, instead of formulating regression methodologies or auto-encoder structures that make use of self-supervision~\cite{tewari2017self,genova2018unsupervised,tran2018nonlinear}, we revisit the optimization-based 3DMM fitting approach by the supervision of deep identity features and by using Generative Adversarial Networks (GANs) as our statistical parametric representation of the facial texture.

In particular, the novelties that this paper brings are:
\begin{itemize}
    \item We show for the first time, to the best of our knowledge, that a large-scale high-resolution statistical reconstruction of the complete facial surface on an unwrapped UV space can be successfully used for reconstruction of arbitrary facial textures even captured in unconstrained recording conditions\footnote{In the very recent works, it was shown that it is feasible to reconstruct the non-visible parts a UV space for facial texture completion\cite{deng2017uv} and that GANs can be used to generate novel high-resolution faces\cite{slossberg2018high}. Nevertheless, our work is the first one that demonstrates that a GAN can be used as powerful statistical texture prior and reconstruct the complete texture of arbitrary facial images.}. 
  \item We formulate a novel 3DMM fitting strategy which is based on GANs and a differentiable renderer. 
  \item We devise a novel cost function which combines various content losses on deep identity features from a face recognition network.
  \item We demonstrate excellent facial shape and texture reconstructions in arbitrary recording conditions that are shown to be both photorealistic and identity preserving in qualitative and quantitative experiments. 
 \end{itemize}

\section{History of 3DMM Fitting}

\begin{figure*}
\begin{center}
\includegraphics[width=1.\linewidth]{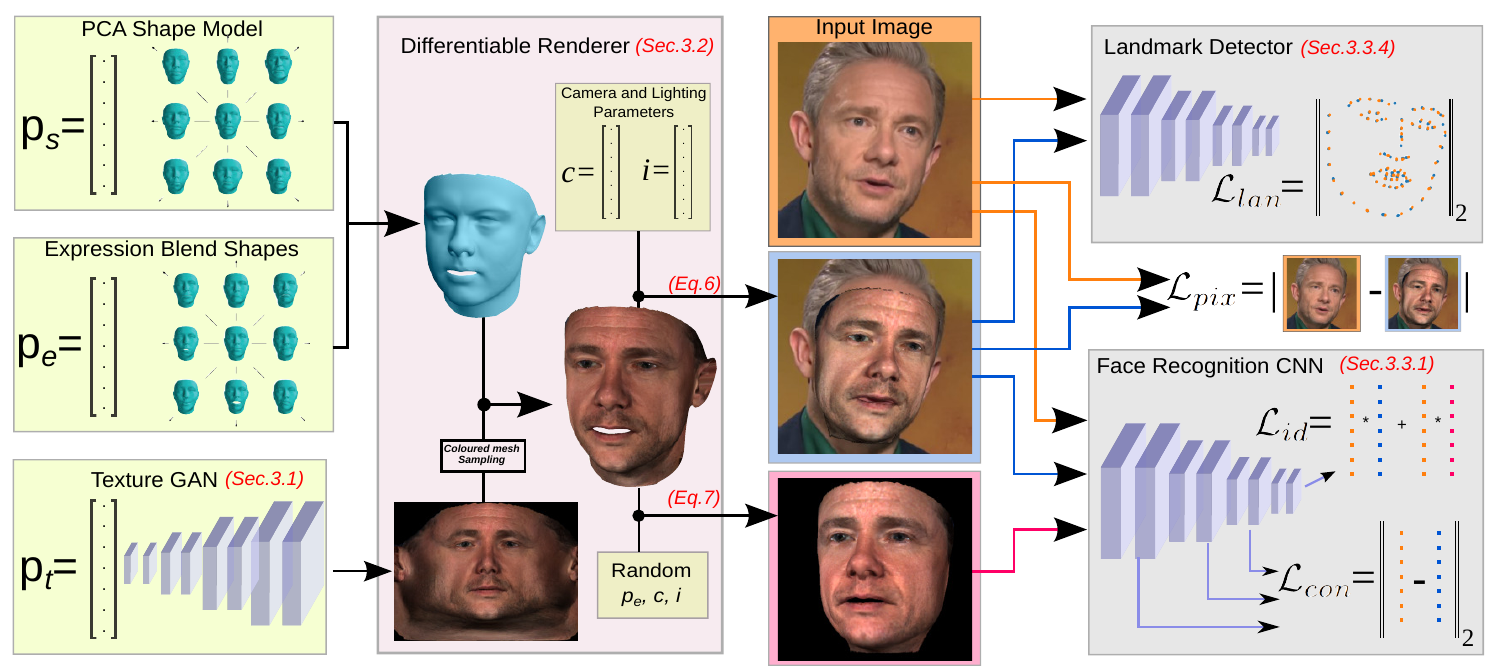}
\vspace{-0.8cm}
\end{center}
  \caption{Detailed overview of the proposed approach. A 3D face reconstruction is rendered by a differentiable renderer (shown in purple). Cost functions are mainly formulated by means of identity features on a pretrained face recognition network (shown in gray) and they are optimized by flowing the error all the way back to the latent parameters ($p_s, p_e, p_t, c, i$, shown in green) with gradient descent optimization. End-to-end differentiable architecture enables us to use computationally cheap and reliable  first order derivatives for optimization thus making it possible to employ deep networks as a generator (i.e,. statistical model) or as a cost function.}
\label{fig:overview}
\end{figure*}

Our methodology naturally extends and generalizes the ideas of texture and shape 3DMM using modern methods for representing texture using GANs, as well as defines loss functions using differentiable renderers and very powerful publicly available face recognition networks~\cite{deng2018arcface}. Before we define our cost function, we will briefly outline the history of 3DMM representation and fitting. 

\subsection{3DMM representation}

The first step is to establish dense correspondences between the training 3D facial meshes and a chosen template with fixed topology in terms of vertices and triangulation.

\subsubsection{Texture}
\label{sec:history_texture}
Traditionally 3DMMs use a UV map for representing texture. UV maps help us to assign 3D texture data into 2D planes with universal per-pixel alignment for all textures. A commonly used UV map is built by cylindrical unwrapping the mean shape into a 2D flat space formulation, which we use to   create an RGB image $\mathbf{I}_{UV}$. Each vertex in the 3D space has a texture coordinate $t_{coord}$ in the UV image plane in which the texture information is stored. A universal function exists, where for each vertex we can sample the texture information from the UV space as $\mathbf{T} = \mathcal{P}(\mathbf{I}_{UV},t_{coord})$.

In order to define a statistical texture representation, all the training texture UV maps are vectorized and Principal Component Analysis (PCA) is applied. Under this model any test texture $\mathbf{T}^0$ is approximated as a linear combination of the mean texture $\mathbf{m}_t$ and a set of bases $\mathbf{U}_t$ as follows:
\begin{equation}
\mathbf{T}(\mathbf{p}_t) \approx \mathbf{m}_t + \mathbf{U}_t \mathbf{p}_t
\label{eq:pca_texture}
\end{equation}
where $\mathbf{p}_{t}$ is the texture parameters for the text sample $\mathbf{T}^0$. 
In the early 3DMM studies, the statistical model of the texture was built with few faces captured in strictly controlled conditions and was used to reconstruct the test albedo of the face. Since, such texture models can hardly represent faces captured in uncontrolled recording conditions (in-the-wild). Recently it was proposed to use statistical models of hand-crafted features such as SIFT or HoG~\cite{booth20173d} directly from in-the-wild faces. The interested reader is referred to \cite{blanz2003face,romdhani2002face} for more details on texture models used in 3DMM fitting algorithms. 

The recent 3D face fitting methods~\cite{tewari2017self,tran2018nonlinear,garrido2016reconstruction} still make use of similar statistical models for the texture. Hence, they can naturally represent only the low-frequency components of the facial texture (please see   Fig.~\ref{fig:comparison}).

\subsubsection{Shape}
The method of choice for building statistical models of facial or head 3D shapes is still PCA~\cite{jolliffe2011principal}. Assuming that the 3D shapes in correspondence comprise of $N$ vertexes, i.e. $\mathbf{s} = {\left[\mathbf{x}_1^\mathsf{T}, \ldots, \mathbf{x}_N^\mathsf{T}\right]}^\mathsf{T} = {\left[x_1, y_1, z_1, \ldots, x_N, y_N, z_N\right]}^\mathsf{T}$. In order to represent both variations in terms of identity and expression, generally two linear models are used. The first is learned from facial scans displaying the neutral expression (i.e., representing identity variations) and the second is learned from displacement vectors (i.e., representing expression variations). Then a test facial shape $\mathbf{S}(\mathbf{p}_{s,e})$ can be written as
\begin{equation}
\mathbf{S}(\mathbf{p}_{s,e}) \approx \mathbf{m}_{s,e} + \mathbf{U}_{s,e} \mathbf{p}_{s,e}
\label{eq:pca_shape}
\end{equation}
where $\mathbf{m}_{s,e}$ in the mean shape vector, $\mathbf{U}_{s,e} \in \mathbb{R}^{3N\times n_{s,e}}$ is $\mathbf{U}_{s,e} = [\mathbf{U}_s, \mathbf{U}_e]$ where the $\mathbf{U}_s$ are the bases that correspond to identity variations, and $\mathbf{U}_e$ the bases that correspond to expression. Finally, $\mathbf{p}_{s,e}$ are the $n_{s,e}$ shape parameters which can be split accordingly to the identity and expression bases: $\mathbf{p}_{s,e}$ = [$\mathbf{p_s}$, $\mathbf{p_e}$].

\subsection{Fitting}

3D face and texture reconstruction by fitting a 3DMM is performed by solving a non-linear energy based cost optimization problem that recovers a set of parameters $\mathbf{p} = [\mathbf{p}_{s,e}, \mathbf{p}_t, \mathbf{p}_{c}, \mathbf{p}_{l}]$ where $\mathbf{p}_c$ are the parameters related to a camera model and $\mathbf{p}_{l}$ are the parameters related to an illumination model. The optimization can be formulated as: 
\begin{equation}
    \min_{\mathbf{p}} \mathcal{E}(\mathbf{p}) = ||\mathbf{I}^0(\mathbf{p}) - \mathbf{W}(\mathbf{p})||_2^2 + Reg(\{\mathbf{p}_{s,e},\mathbf{p}_t\})
\end{equation}
where $\mathbf{I}^0$ is the test image to be fitted and $\mathbf{W}$ is a vector produced by a physical image formation process (i.e., rendering) controlled by $\mathbf{p}$. Finally, $Reg$ is the regularization term that is mainly related to texture and shape parameters.

Various methods have been proposed for numerical optimization of the above cost functions~\cite{hu2017efficient,bas2016fitting}. A notable recent approach is \cite{booth20173d} which uses handcrafted features (i.e., $\mathbf{H}$) for texture representation simplified the cost function as:     
\begin{equation}
\!\!\!\min_{\mathbf{p}^r} \mathcal{E}(\mathbf{p}^r)\! =\! || \mathbf{H}(\mathbf{I}^0(\mathbf{p}^r)) - \mathbf{H}(\mathbf{W}(\mathbf{p}^r))||_{\mathbf{A}}^2 \!+\! Reg(\mathbf{p}_{s,e})
\label{eq:james_energy}
\end{equation}
where $||\mathbf{a}||_{\mathbf{A}}^2 = \mathbf{a}^T\mathbf{A}\mathbf{a}$, $\mathbf{A}$ is the orthogonal space to the statistical model of the texture and  $\mathbf{p}^r$ is the set of reduced parameters $\mathbf{p}^r = \{\mathbf{p}_{s,e}, \mathbf{p}_{c}\}$. The optimization problem in Eq.~\ref{eq:james_energy} is solved by Gauss-Newton method. The main drawback of this method is that the facial texture in not reconstructed. 



In this paper, we generalize the 3DMM fittings and introduce the following novelties:
\begin{itemize}
    \item We use a GAN on high-resolution UV maps as our statistical representation of the facial texture. That way we can reconstruct textures with high-frequency details.  
    
    \item Instead of other cost functions used in the literature such as low-level $\ell_1$ or $\ell_2$ loss (e.g., RGB values~\cite{piotraschke2016automated}, edges~\cite{romdhani2005estimating}) or hand-crafted features (e.g., SIFT~\cite{booth20173d}), we propose a novel cost function that is based on feature loss from the various layers of publicly available face recognition embedding network~\cite{deng2018arcface}. Unlike others, deep identity features are very powerful at preserving identity characteristics of the input image.
    
    \item We replace physical image formation stage with a differentiable renderer to make use of first order derivatives (i.e., gradient descent). Unlike its alternatives, gradient descent provides computationally cheaper and more reliable derivatives through such deep architectures (i.e., above-mentioned texture GAN and identity DCNN).
     
\end{itemize}

\section{Approach}

We propose an optimization-based 3D face reconstruction approach from a single image that employs a high fidelity texture generation network as statistical prior as illustrated in Fig.~\ref{fig:overview}. To this end, the reconstruction mesh is formed by 3D morphable shape model; textured by the generator network's output UV map; and projected into 2D image by a differentiable renderer. The distance between the rendered image and the input image is minimized in terms of a number of cost functions by updating the latent parameters of 3DMM and the texture network with gradient descent. We mainly formulate these functions based on rich features of face recognition network~\cite{deng2018arcface,schroff2015facenet,parkhi2015deep} for smoother convergence and landmark detection network~\cite{deng2018cascade} for alignment and rough shape estimation. 

The following sections introduce firstly our novel texture model that employs a generator network trained by progressive growing GAN framework. After describing the procedure for image formation with differentiable renderer, we formulate our cost functions and the procedure for fitting our shape and texture models onto a test image.

\subsection{GAN Texture Model}
Although conventional PCA is powerful enough to build a decent shape and texture model, it is often unable to capture high frequency details and ends up having blurry textures due to its Gaussian nature. This becomes more apparent in texture modelling which is a key component in 3D reconstruction to preserve identity as well as photo-realism.

GANs are shown to be very effective at capturing such details. However, they suffer from preserving 3D coherency~\cite{goodfellow2016nips} of the target distribution when the training images are semi-aligned. We found that a GAN trained with UV representation of real textures with per pixel alignment avoids this problem and is able to generate realistic and coherent UVs from $99.9\%$ of its latent space while at the same time  generalizing well to unseen data.

In order to take advantage of this perfect harmony, we train a progressive growing GAN~\cite{karras2018progressive} to model distribution of UV representations of 10,000 high resolution textures and use the trained generator network 
\begin{equation}
\mathcal{G}(\mathbf{p}_t): \mathbb{R}^{512} \rightarrow \mathbb{R}^{H\times W \times C}
\label{eq:gan}
\end{equation}
as texture model that replaces 3DMM texture model in Eq.~\ref{eq:pca_texture}.

While fitting with linear models, i.e. 3DMM, is as simple as linear transformation, fitting with a generator network can be formulated as an optimization that minimizes per-pixel Manhattan distance between target texture in UV space $\mathbf{I}_{uv}$ and the network output $\mathcal{G}(\mathbf{p}_t)$ with respect to the latent parameter $\mathbf{p}_t$, i.e. $\min_{\mathbf{p}_t} |\mathcal{G}(\mathbf{p}_t)-\mathbf{I}_{uv}|$.

\subsection{Differentiable Renderer}
Following \cite{genova2018unsupervised}, we employ a differentiable renderer to project 3D reconstruction into a 2D image plane based on deferred shading model with given camera and illumination parameters. Since color and normal attributes at each vertex are interpolated at the corresponding pixels with barycentric coordinates, gradients can be easily backpropagated through the renderer to the latent parameters.

A 3D textured mesh at the center of Cartesian origin $[0,0,0]$ is projected onto 2D image plane by a pinhole camera model with the camera standing at $[x_c,y_c,z_c]$, directed towards $[x_c',y_c',z_c']$ and with the focal length $f_c$. The illumination is modelled by phong shading given 1) direct light source at 3D coordinates $[x_l,y_l,z_l]$ with color values $[r_l,g_l,b_l]$, and 2) color of ambient lighting $[r_a,g_a,b_a]$.

Finally, we denote the rendered image given geometry ($\mathbf{p}_{s,e}$), texture ($\mathbf{p}_{t}$), camera ($\mathbf{p}_c=[x_c,y_c,z_c,x_c',y_c',z_c',f_c]$) and lighting parameters ($\mathbf{p}_l = [x_l,y_l,z_l,r_l,g_l,b_l,r_a,g_a,b_a]$ by the following:
\begin{equation}
\mathbf{I}^{\mathcal{R}} = \mathcal{R}(\mathbf{S}(\mathbf{p}_s,\mathbf{p}_e), \mathcal{P}( \mathcal{G}(\mathbf{p}_t)),\mathbf{p}_c,\mathbf{p}_l)
\end{equation}
where we construct shape mesh by 3DMM as given in Eq.~\ref{eq:pca_shape} and texture by GAN generator network as in Eq.~\ref{eq:gan}. Since our differentiable renderer supports only color vectors, we sample from our generated UV map to get vectorized color representation as explained in Sec.~\ref{sec:history_texture}.

Additionally, we render a secondary image with random expression, pose and illumination in order to generalize identity related parameters well with those variations. We sample expression parameters from a normal distribution as $\hat{\mathbf{p}_e} \sim \mathcal{N}(\mu=0,\sigma=0.5)$ and sample camera and illumination parameters from the Gaussian distribution of 300W-3D dataset as $\hat{\mathbf{p}}_c \sim \mathcal{N}(\hat{\mu_c},\hat{\sigma_c})$ and $\hat{\mathbf{p}_l} \sim \mathcal{N}(\hat{\mu_l},\hat{\sigma_l})$. This rendered image of the same identity as $\mathbf{I}^{\mathcal{R}}$ (i.e., with same $\mathbf{p}_s$ and $\mathbf{p}_t$ parameters) is expressed by the following:
\begin{equation}
\hat{\mathbf{I}}^{\mathcal{R}} = \mathcal{R}(\mathbf{S}(\mathbf{p}_s,\hat{\mathbf{p}_e}),\mathcal{P}(\mathcal{G}(\mathbf{p}_t)),\hat{\mathbf{p}_c},\hat{\mathbf{p}_l})
\end{equation}

\subsection{Cost Functions}
Given an input image $\mathbf{I}^0$, we optimize all of the aforementioned parameters simultaneously with gradient descent updates. In each iteration, we simply calculate the forthcoming cost terms for the current state of the 3D reconstruction, and take the derivative of the weighted error with respect to the parameters using backpropagation.

\subsubsection{Identity Loss}
With the availability of large scale datasets, CNNs have shown incredible performance on many face recognition benchmarks. Their strong identity features are robust to many variations including pose, expression, illumination, age etc. These features are shown to be quite effective at many other tasks including novel identity synthesizing~\cite{gecer2018facegan}, face normalization~\cite{cole2017synthesizing} and  3D face reconstruction~\cite{genova2018unsupervised}. In our approach, we take advantage of an off-the-shelf state-of-the-art face recognition network~\cite{deng2018arcface}\footnote{We empirically deduced that other face recognition networks work almost equally well and this choice is orthogonal to the proposed approach.} in order to capture identity related features of an input face image and optimize the latent parameters accordingly. More specifically, given a pretrained face recognition network $\mathcal{F}^n(\mathbf{I}): \mathbb{R}^{H\times W \times C} \rightarrow \mathbb{R}^{512}$ consisting of $n$ convolutional filters, we calculate the cosine distance between the identity features (i.e., embeddings) of the real target image and our rendered images as following:
\begin{equation}
\mathcal{L}_{id} = 1 - \frac{\mathcal{F}^n(\mathbf{I}^0) . \mathcal{F}^n( \mathbf{I}^\mathcal{R})}{||\mathcal{F}^n(\mathbf{I}^0)||_2 ||\mathcal{F}^n(\mathbf{I}^\mathcal{R})||_2}
\label{eq:id_loss}
\end{equation}
We formulate  an additional identity loss on the rendered image $\hat{\mathbf{I}}^\mathcal{R}$ that is rendered with random pose, expression and lighting. This loss ensures that our reconstruction resembles the target identity under different conditions. We formulate it by replacing $\mathbf{I}^\mathcal{R}$ by $\hat{\mathbf{I}}^\mathcal{R}$ in Eq.~\ref{eq:id_loss} and it is denoted as $\hat{\mathcal{L}}_{id}$.

\begin{figure*}[ht]
\begin{center}
\includegraphics[width=1.\linewidth]{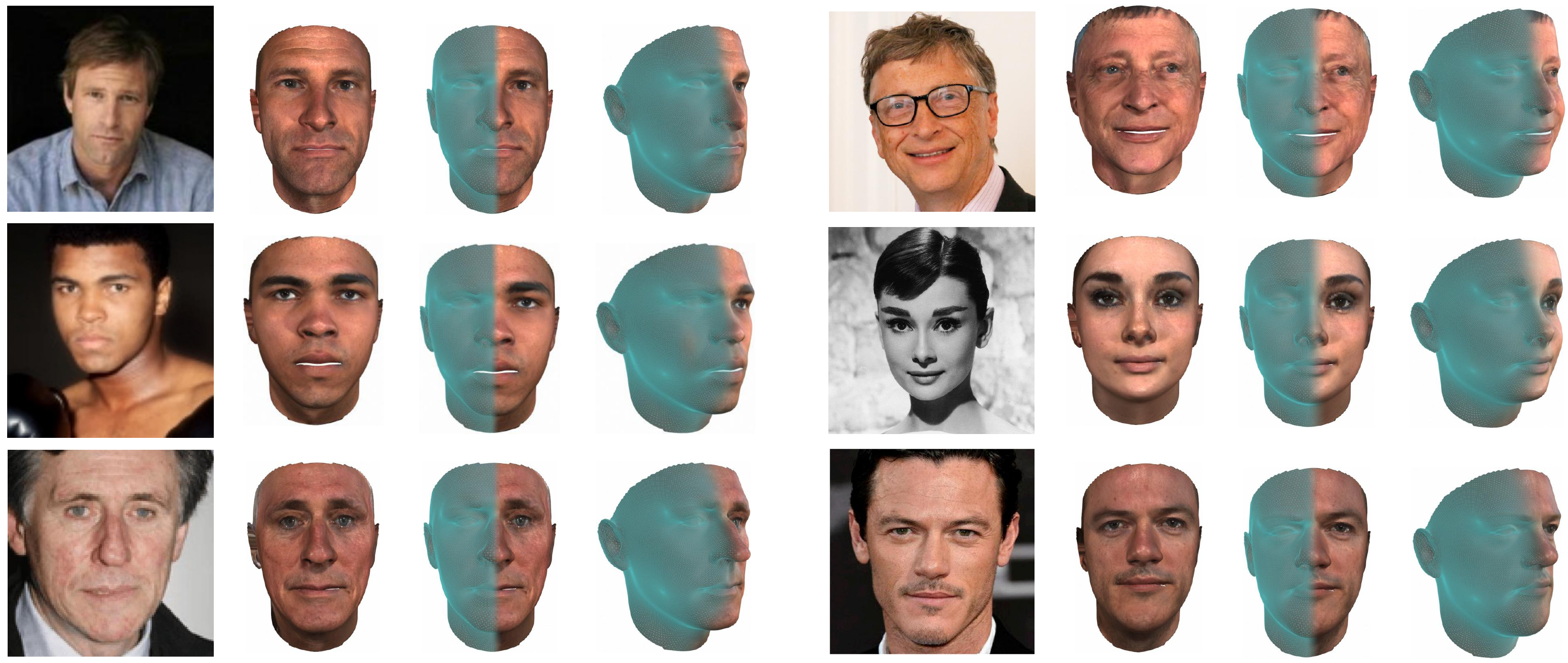}
\vspace{-0.5cm}
\caption{Example fits of our approach for the images from various datasets. Please note that our fitting approach is robust to occlusion (e.g., glasses), low resolution and black-white in the photos and generalizes well with ethnicity, gender and age. The reconstructed textures are very well at capturing high frequency details of the identities; likewise, the reconstructed geometries from 3DMM are surprisingly good at identity preservation thanks to the identity features used, e.g. crooked nose at bottom-left, dull eyes at bottom-right and chin dimple at top-left}
\vspace{-0.5cm}
\label{fig:qualitative}
\end{center}
\end{figure*}

\subsubsection{Content Loss}
Face recognition networks are trained to remove all kinds of attributes (e.g. expression, illumination, age, pose) other than abstract identity information throughout the convolutional layers. Despite their strength, the activations in the very last layer discard some of the mid-level features that are useful for 3D reconstruction, e.g. variations that depend on age. Therefore we found it effective to accompany identity loss by leveraging intermediate representations in the face recognition network that are still robust to pixel-level deformations and not too abstract to miss some details. To this end, normalized euclidean distance of intermediate activations, namely content loss, is minimized between input and rendered image with the following loss term:
\begin{equation}
    \mathcal{L}_{con} =  \sum_j^n{\dfrac{||\mathcal{F}^j(\mathbf{I}^0) - \mathcal{F}^j(\mathbf{I}^{\mathcal{R}})||_2}{H_{\mathcal{F}^j} \times W_{\mathcal{F}^j} \times C_{\mathcal{F}^j}}}
\end{equation}

\subsubsection{Pixel Loss}
While identity and content loss terms optimize albedo of the visible texture, lighting conditions are optimized based on pixel value difference directly. While this cost function is relatively primitive, it is sufficient to optimize lighting parameters such as ambient colors, direction, distance and color of a light source. We found that optimizing illumination parameters jointly with others helped to improve albedo of the recovered texture. Furthermore, pixel loss support identity and content loss with fine-grained texture as it supports highest available resolution while images needs to be downscaled to $112 \times 112$ before identity and content loss. The pixel loss is defined by pixel level $\ell_1$ loss function as:
\begin{equation}
    \mathcal{L}_{pix} =  ||\mathbf{I}^0 - \mathbf{I}^{\mathcal{R}}||_1 
\end{equation}

\subsubsection{Landmark Loss}
The face recognition network $\mathcal{F}$ is pre-trained by the images that are aligned by similarity transformation to a fixed landmark template. To be compatible with the network, we align the input and rendered images under the same settings. However, this process disregards the aspect ratio and scale of the reconstruction. Therefore, we employ a deep face alignment network~\cite{deng2018cascade} $\mathcal{M}(\mathbf{I}): \mathbb{R}^{H\times W \times C} \rightarrow \mathbb{R}^{68 \times 2}$ to detect landmark locations of the input image and align the rendered geometry onto it by updating the shape, expression and camera parameters. That is, camera parameters are optimized to align with the pose of image $\mathbf{I}$ and geometry parameters are optimized for the rough shape estimation. As a natural consequence, this alignment drastically improves the effectiveness of the pixel and content loss, which are sensitive to misalignment between the two images.

The alignment error is achieved by point-to-point euclidean distances between detected landmark locations of the input image and 2D projection of the 3D reconstruction landmark locations that is available as meta-data of the shape model. Since landmark locations of the reconstruction heavily depend on camera parameters, this loss is great a source of information the alignment of the reconstruction onto input image and  is formulated as  following:
\begin{equation}
    \mathcal{L}_{lan} = ||\mathcal{M}(\mathbf{I}^0) -  \mathcal{M}(\mathbf{I}^{\mathcal{R}})||_2
\end{equation}

\subsection{Model Fitting}
We first roughly align our reconstruction  to the input image by optimizing shape, expression and camera parameters by:
$\min_{\mathbf{p}^r} \mathcal{E}(\mathbf{p}^r) = \lambda_{lan}\mathcal{L}_{lan}$.
We then simultaneously optimize all of our parameters with gradient descent and backpropagation so as to minimize weighted combination of above loss terms in the following:
\begin{equation}
\begin{aligned}
\!\!\!\min_{\mathbf{p}} \mathcal{E}(\mathbf{p}) = \lambda_{id}\mathcal{L}_{id} \!+ \hat{\lambda}_{id}\hat{\mathcal{L}}_{id} \!+ \lambda_{con}\mathcal{L}_{con}  +\! \lambda_{pix}\mathcal{L}_{pix} \\+ \lambda_{lan}\mathcal{L}_{lan} + \lambda_{reg} Reg(\{\mathbf{p}_{s,e},\mathbf{p}_l\})
\end{aligned}
\end{equation}
where we weight each of our loss terms with $\lambda$ parameters. 
In order to prevent our shape and expression models and lighting parameters from exaggeration   to arbitrarily bias our loss terms, we regularize those parameters by $Reg(\{\mathbf{p}_{s,e},\mathbf{p}_l\})$.

\def \myvar {0.95cm} 
\begin{figure*}[th!]
\begin{tabular}{cl}
\vspace{0.4cm} & \multirow{11}{*}{\includegraphics[width=0.75\linewidth]{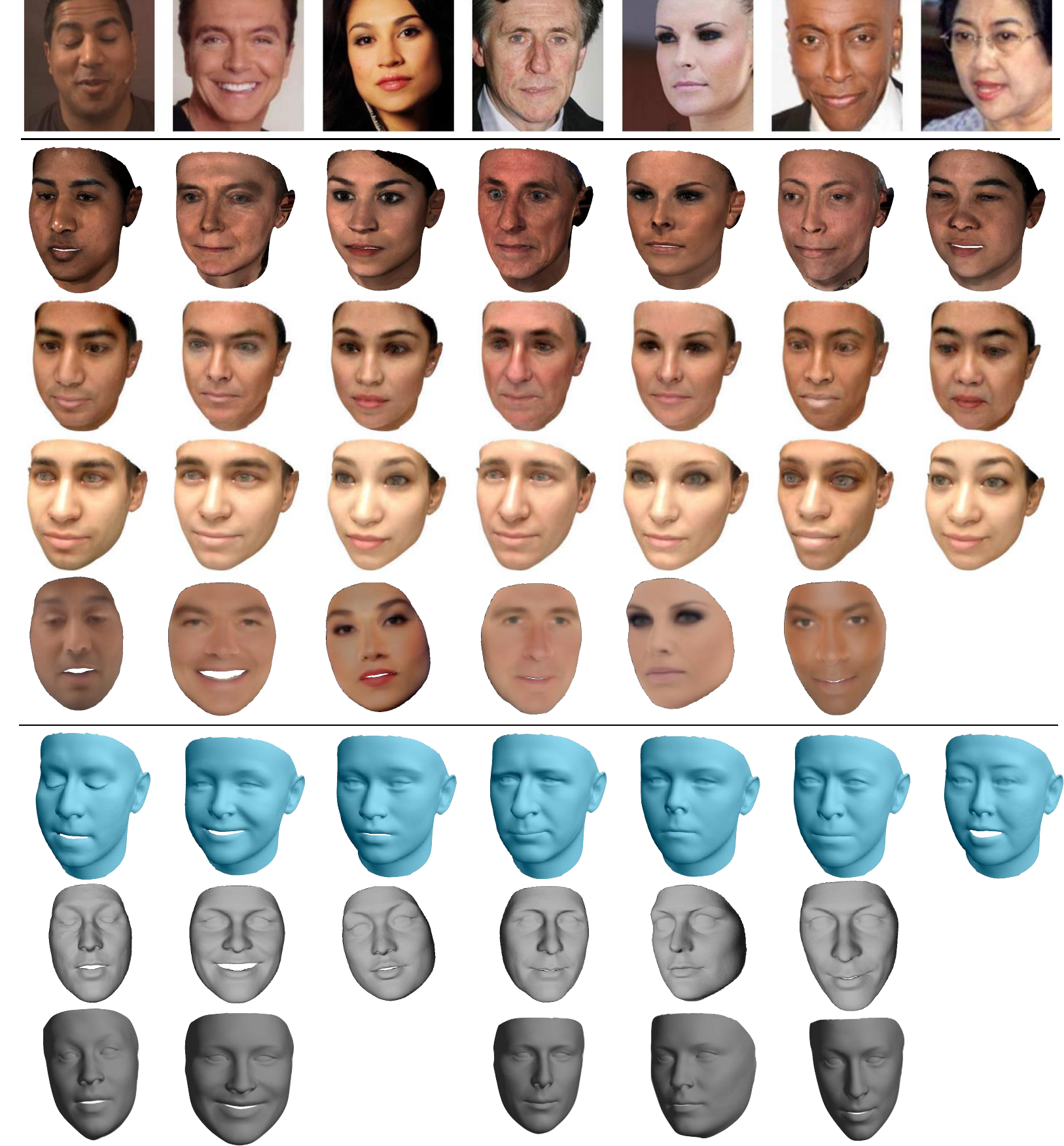}} \\ 
Input Images\vspace{1.1cm}\\\textbf{Ours}\\\vspace{\myvar} &                   \\
Genova\\\cite{genova2018unsupervised} \vspace{\myvar} &                   \\
A.T.Tran et al.\\\cite{tran2017regressing} \vspace{\myvar}  &                   \\
Tewari et al.\\\cite{tewari2017self}  \vspace{\myvar}    &                  \\
\textbf{Ours}\\\textbf{Geometry}\vspace{\myvar} &                  \\
Tewari et al.\\\cite{tewari2017self} \vspace{\myvar} &                  \\
L. Tran et al.\\\cite{tran2018nonlinear} \vspace{0.5cm} &                                            
\end{tabular}\vspace{-0.5cm}
  \caption{Comparison of our qualitative results with other state-of-the-art methods in MoFA-Test dataset. Rows 2-5 show comparison with textured geometry and rows 6-8 compare only shapes. The Figure is best viewed in colored and under zoom.}\vspace{-0.5cm}
\label{fig:comparison}
\end{figure*}

\paragraph{Fitting with Multiple Images (\ie Video):}
While the proposed approach can fit a 3D reconstruction from a single image, one can take advantage of more images effectively when available, e.g. from a video recording. This often helps to improve reconstruction quality under challenging conditions, e.g. outdoor, low resolution. While state-of-the-art methods follow naive approaches by averaging either the reconstruction~\cite{tran2017regressing} or features-to-be-regressed~\cite{genova2018unsupervised} before making a reconstruction, we utilize the power of iterative optimization by averaging identity reconstruction parameters ($\mathbf{p}_s, \mathbf{p}_t$) after every iteration. For an image set $\mathbf{I} = \{\mathbf{I}^0,\mathbf{I}^1, \dots,\mathbf{I}^i,\dots, \mathbf{I}^{n_i}\}$, we reformulate our parameters as $\mathbf{p} = [\mathbf{p}_s, \mathbf{p}_e^i, \mathbf{p}_t, \mathbf{p}_c^i, \mathbf{p}_l^i]$ in which we average shape and texture parameters by the following:
\begin{equation}
    \mathbf{p}_s = \sum_{i}^n \mathbf{p}_s^i, \mathbf{p}_t = \sum_{i}^n \mathbf{p}_t^i
    \label{eq:multi-image}
\end{equation}

\section{Experiments}
This section demonstrates the excellent performance of the proposed approach for 3D face reconstruction and shape recovery. We verify this by qualitative results in Figures~\ref{fig:cover}, \ref{fig:qualitative}, qualitative comparisons with the state-of-the-art in Sec.~\ref{sec:exp_comp} and quantitative shape reconstruction experiment on a database with ground truth in Sec.~\ref{sec:exp_micc}.

\subsection{Implementation Details}
For all of our experiments, a given face image is aligned to our fixed template using 68 landmark locations detected by an hourglass 2D landmark detection~\cite{deng2018cascade}. For the identity features, we employ ArcFace~\cite{deng2018arcface} network's pretrained models. For the generator network $\mathcal{G}$, we train a progressive growing GAN~\cite{karras2018progressive} with around 10,000 UV maps from \cite{booth20163d} at the resolution of $512 \times 512$. We use the Large Scale Face Model~\cite{booth20163d} for 3DMM shape model with $n_s=158$ and the expression model learned from 4DFAB database~\cite{cheng20174dfab} with $n_e=29$. During fitting process, we optimize parameters using Adam Solver~\cite{kingma2014adam} with 0.01 learning rate. And we set our balancing factors as the following: $\lambda_{id}:2.0, \hat{\lambda}_{id}:2.0, \lambda_{con}:50.0, \lambda_{pix}:1.0, \lambda_{lan}:0.001, \lambda_{reg}: \{0.05, 0.01\}$. The Fitting converges in around 30 seconds on an Nvidia GTX 1080 TI GPU for a single image.

\subsection{Qualitative Comparison to the State-of-the-art}
\label{sec:exp_comp}
Fig.~\ref{fig:comparison} compares our results with the most recent face reconstruction studies~\cite{tewari2017mofa,tewari2017self,genova2018unsupervised,tran2017regressing,tran2018nonlinear} on a subset of MoFA test-set. The first four rows after input images show a comparison of our shape and texture reconstructions to \cite{genova2018unsupervised,tran2017regressing,tewari2017self} and the last three rows show our reconstructed geometries without texture compared to \cite{tewari2017self,tran2018nonlinear}. All in all, our method outshines all others with its high fidelity photorealistic texture reconstructions. Both of our texture and shape reconstructions manifest strong identity characteristics of the corresponding input images from the thickness and shape of the eyebrows to wrinkles around the mouth and forehead. 


\begin{table}[t]
\setlength{\tabcolsep}{4pt}
\centering
\begin{tabular}{|l|cccccc|}
\hline
 & \multicolumn{2}{|c|}{\emph{Cooperative}}  & \multicolumn{2}{|c|}{\emph{Indoor}} & \multicolumn{2}{|c|}{\emph{Outdoor}}\\
 \emph{Method} & \multicolumn{2}{|c|}{Mean \ \ Std.}  & \multicolumn{2}{|c|}{Mean \ \ Std.} & \multicolumn{2}{|c|}{Mean \ \ Std.}\\
 \hline \hline
Tran \etal~\cite{tran2017regressing} & 1.93 & 0.27 & 2.02 & 0.25 & 1.86 & 0.23 \\
Booth \etal~\cite{booth20173d} & 1.82 & 0.29 & 1.85 & 0.22 & 1.63 & 0.16 \\
Genova \etal~\cite{genova2018unsupervised} & 1.50 & 0.13 & 1.50 & 0.11 & 1.48 & 0.11 \\
\emph{Ours} & \textbf{0.95} & \textbf{0.107} & \textbf{0.94} & \textbf{0.106} & \textbf{0.94} & \textbf{0.106}\\
\hline
\end{tabular}
\caption{Accuracy results for the meshes on the MICC Dataset using point-to-plane distance. The table reports the mean error (Mean), the standard deviation (Std.).}
\vspace{-0.4cm}
\label{tab:micc_results}
\end{table}

\subsection{3D shape recovery on MICC dataset}
\label{sec:exp_micc}
We evaluate the shape reconstruction performance of our method on MICC Florence 3D Faces dataset (MICC)~\cite{bagdanov2011florence} in Table~\ref{tab:micc_results}. The dataset provides 3D scans of 53 subjects as well as their short video footages under three difficulty settings: 'cooperative', 'indoor' and 'outdoor'. Unlike \cite{genova2018unsupervised,tran2017regressing} which processes all the frames in a video, we uniformly sample only 5 frames from each video regardless of their zoom level. And, we run our method with multi-image support for these 5 frames for each video separately as shown in Eq.~\ref{eq:multi-image}. Each test mesh is cropped at a radius of $95$\emph{mm} around the tip of the nose according to \cite{tran2017regressing} in order to evaluate the shape recovery of the inner facial mesh. We perform dense alignment between each predicted mesh and its corresponding ground truth mesh, by implementing an iterative closest point (ICP) method~\cite{besl1992method}. As evaluation metric, we follow \cite{genova2018unsupervised} to measure the error by average symetric point-to-plane distance. 

Table~\ref{tab:micc_results} reports the normalized point-to-plain errors in millimeters. It is evident that we have improved the absolute error compared to the other two state-of-the-art methods by $36\%$. Our results are shown to be consistent across all different settings with minimal standard deviation from the mean error.

\subsection{Ablation Study}
Fig.~\ref{fig:ablation} shows an ablation study on our method where the full model reconstructs the input face better than its variants, something that suggests that each of our components significantly contributes towards a good reconstruction. Fig.~\ref{fig:ablation}(c) indicates albedo is well disentangled from illumination and our model capture the light direction accurately.

While Fig.~\ref{fig:ablation}(d-f) shows each of the identity terms contributes to preserve identity, Fig.~\ref{fig:ablation}(h) demonstrates the significance identity features altogether. Still, overall reconstruction utilizes pixel intensities to capture better albedo and illumination as shown in Fig.~\ref{fig:ablation}(g). Finally, Fig.~\ref{fig:ablation}(i) shows the superiority of our textures over PCA-based ones.

\def \obamavar {0.15}
\begin{figure}[t]
 \centering 
\begin{subfigure}{\obamavar\textwidth}
  \includegraphics[width=1.0\textwidth,trim={0 1cm 0 0.5cm},clip]{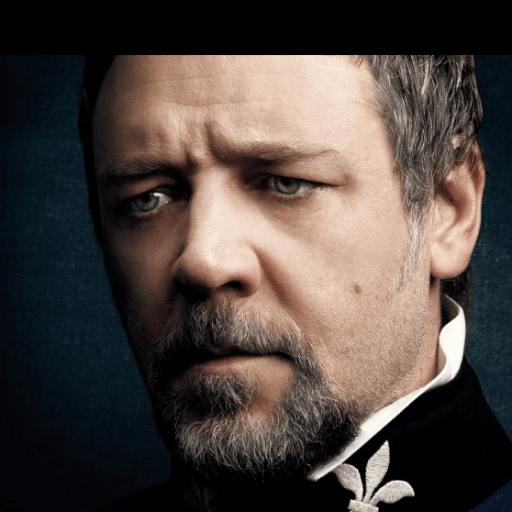}
\tiny\caption{$\mathbf{I^0}$}\end{subfigure}
\begin{subfigure}{\obamavar\textwidth}
  \includegraphics[width=1.0\textwidth,trim={0 1cm 0 0.5cm},clip]{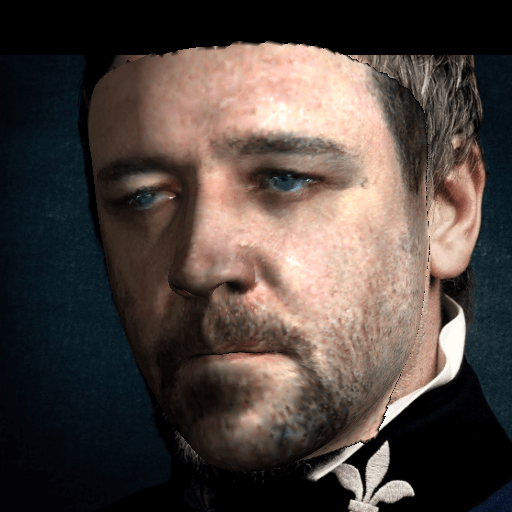}
\tiny\caption{$\mathbf{I^\mathcal{R}}$}\end{subfigure}
\begin{subfigure}{\obamavar\textwidth}
  \includegraphics[width=1.0\textwidth,trim={0 1cm 0 0.5cm},clip]{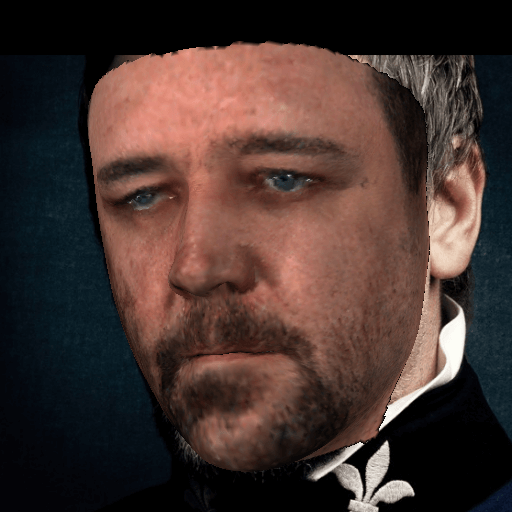}
\tiny\caption{$\mathbf{I^\mathcal{R}}$ albedo}\end{subfigure}
\begin{subfigure}{\obamavar\textwidth}
  \includegraphics[width=1.0\textwidth,trim={0 1cm 0 0.5cm},clip]{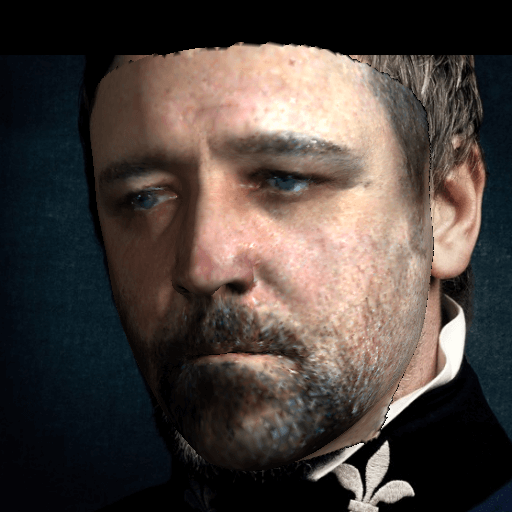}
\tiny\caption{$\mathbf{I^\mathcal{R}} \setminus \mathcal{L}_{id}$}\end{subfigure}
\begin{subfigure}{\obamavar\textwidth}
  \includegraphics[width=1.0\textwidth,trim={0 1cm 0 0.5cm},clip]{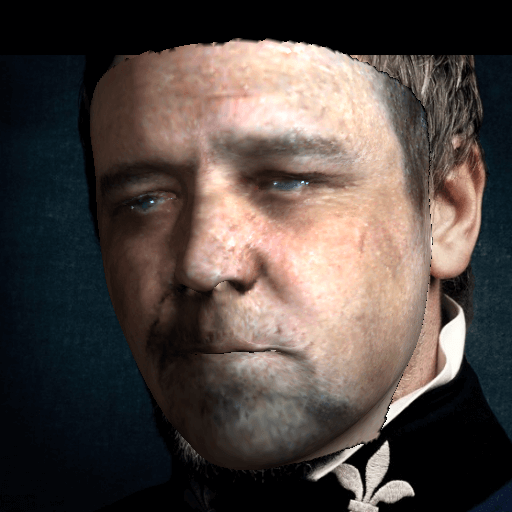}
\tiny\caption{$\mathbf{I^\mathcal{R}} \setminus \hat{\mathcal{L}}_{id}$}\end{subfigure}
\begin{subfigure}{\obamavar\textwidth}
  \includegraphics[width=1.0\textwidth,trim={0 1cm 0 0.5cm},clip]{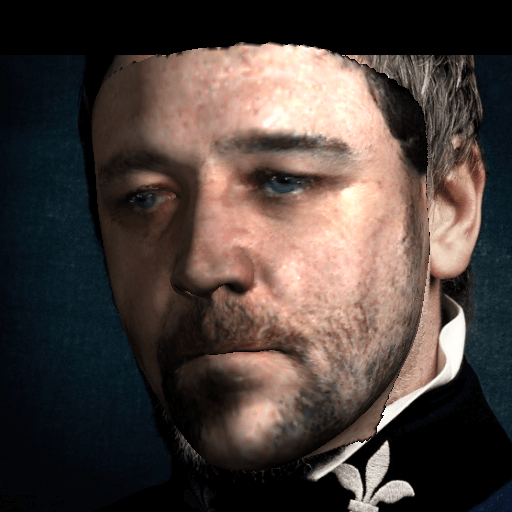}
\tiny\caption{$\mathbf{I^\mathcal{R}} \setminus \mathcal{L}_{con}$}\end{subfigure}
\begin{subfigure}{\obamavar\textwidth}
  \includegraphics[width=1.0\textwidth,trim={0 1cm 0 0.5cm},clip]{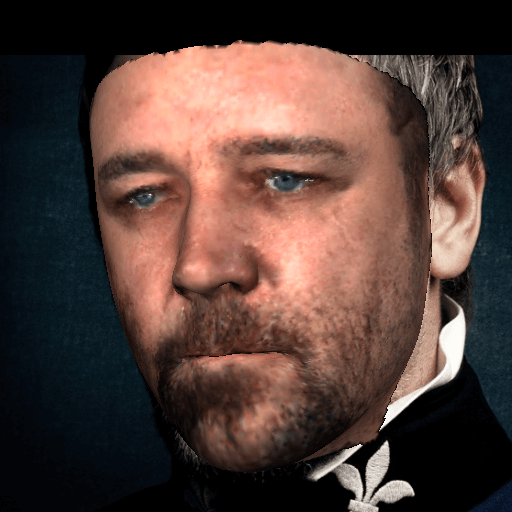}
\tiny\caption{$\mathbf{I^\mathcal{R}} \setminus \mathcal{L}_{pix} $}\end{subfigure}
\begin{subfigure}{\obamavar\textwidth}
  \includegraphics[width=1.0\textwidth,trim={0 1cm 0 0.5cm},clip]{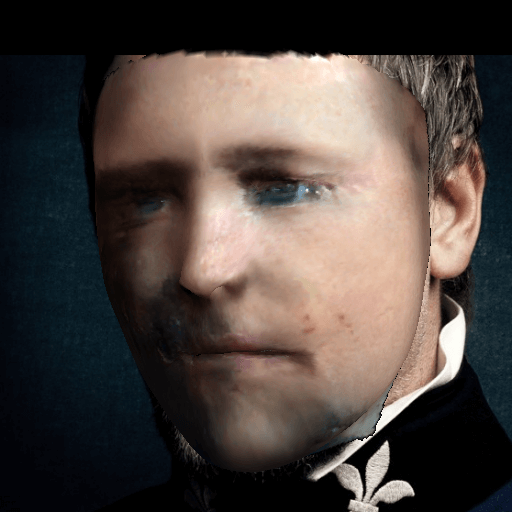}
\tiny\caption{$\!\mathbf{I^{\!\mathcal{R}}}\!\setminus\!\{\!\mathcal{L}_{id}\!,\! \hat{\mathcal{L}}_{id}\!,\!\mathcal{L}_{con}\!\!\}$}\end{subfigure}
\begin{subfigure}{\obamavar\textwidth}
  \includegraphics[width=1.0\textwidth,trim={0 1cm 0 0.5cm},clip]{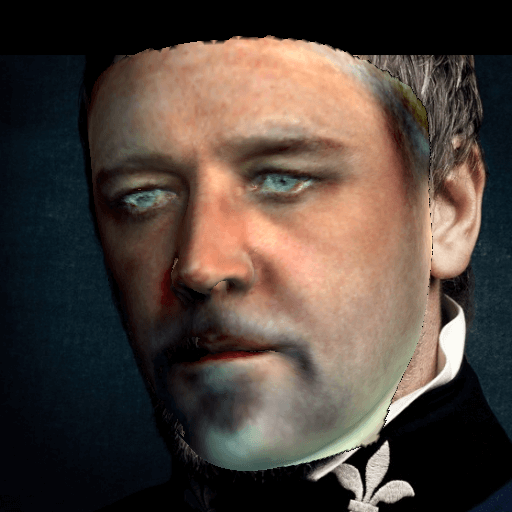}
\tiny\caption{$\mathbf{I^\mathcal{R}}$ with $\mathbf{T}(\mathbf{p_t})$}\end{subfigure}
\caption{Contributions of the components or loss terms of the proposed approach with an leave-one-out ablation study.}\vspace{-0.5cm}
\label{fig:ablation}
\end{figure}

\section{Conclusion}
In this paper, we revisit optimization-based 3D face reconstruction under a new perspective, that is, we utilize the power of recent machine learning techniques such as GANs and face recognition network as statistical texture model and as energy function respectively.

To the best of our knowledge, this is the first time that GANs are used for model fitting and they have shown excellent results for high quality texture reconstruction. The proposed approach shows identity preserving high fidelity 3D reconstructions in qualitative and quantitative experiments.

\paragraph{Acknowledgements:}
Baris Gecer is funded by the Turkish Ministry of
National Education. Stefanos Zafeiriou acknowledges support by EPSRC Fellowship DEFORM (EP/S010203/1) and a Google Faculty Award. 

\clearpage
{\small
\bibliographystyle{ieee}
\bibliography{egbib_arxiv}
}

\clearpage
\begin{appendices}

\section{Experiments on LFW}

In order to evaluate identity preservation capacity of the proposed method, we run two face recognition experiments on Labelled Faces in the Wild (LFW) dataset~\cite{Huang2012a}. Following~\cite{genova2018unsupervised}, we feed real LFW images and rendered images of their 3D reconstruction by our method to a pretrained face recognition network, namely VGG-Face\cite{Parkhi15}. We then compute the activations at the embedding layer and measure cosine similarity between 1) real and rendered images and 2) renderings of same/different pairs. 

\begin{figure}[b]
\begin{center}
\includegraphics[width=1.0\linewidth,trim={1.6cm 0 1.6cm 0},clip]{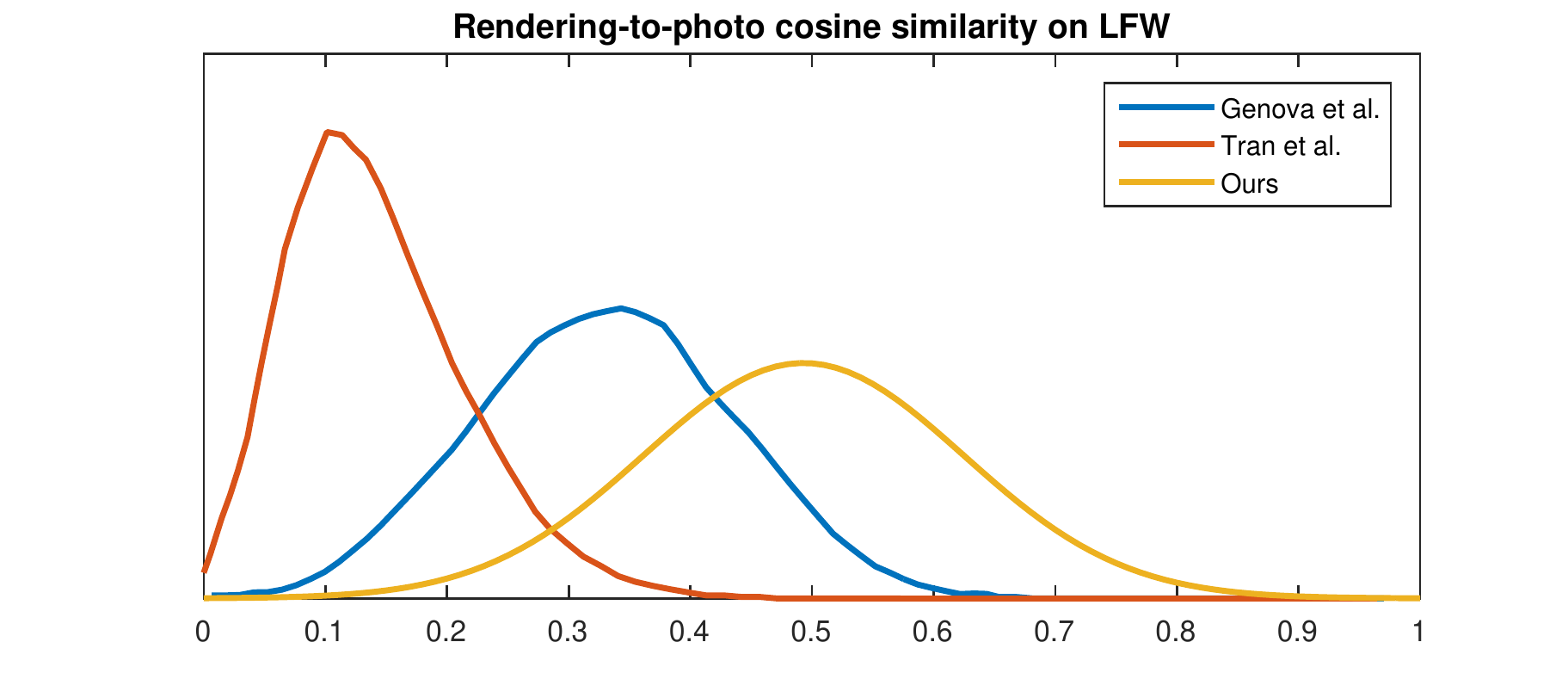}
\end{center}
  \caption{Cosine similarity distributions of rendered and real images LFW based on activations at the embedding layer of VGG-Face network\cite{Parkhi15}. Our method achieves more than 0.5 similarity on average which \cite{genova2018unsupervised} has 0.35 average similarity and \cite{tran2017regressing} 0.16 average similarity. Camera and lighting parameters are fixed for all renderings.}
\label{fig:sim}
\end{figure}

\begin{figure}[b]
\begin{center}
\includegraphics[width=1.0\linewidth,trim={1.6cm 0 1.6cm 0},clip]{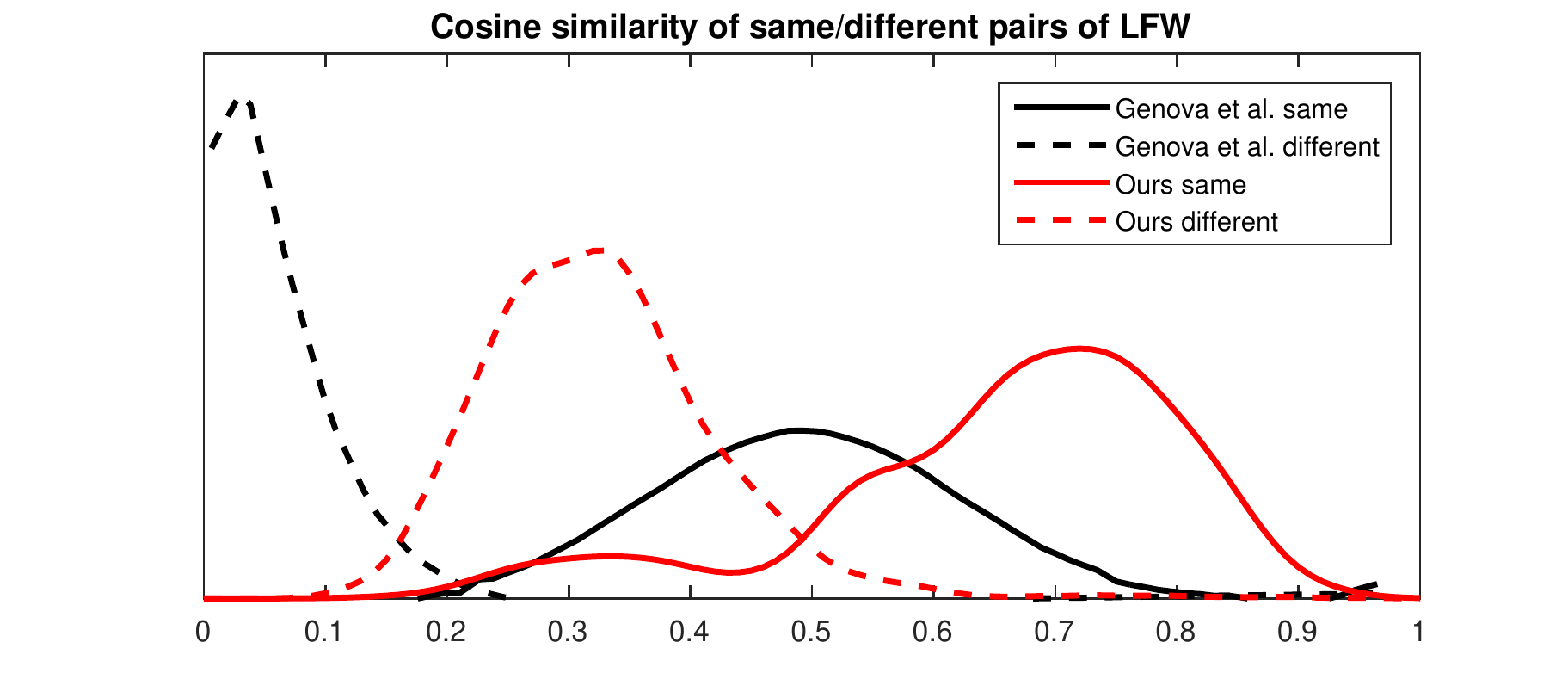}
\end{center}
  \caption{Our method successfully preserve identity so that distribution of cosine similarity of same/different pairs is separable by thresholding. Camera and lighting parameters are fixed for all renderings.}
\label{fig:pairs}
\end{figure}

In Fig. \ref{fig:sim} and \ref{fig:pairs}, we have quantitatively showed that our method is better at identity preservation and photorealism (i.e., as the pretrained network is trained by real images) than other state-of-the-art deep 3D face reconstruction approaches \cite{genova2018unsupervised,tran2017regressing}.

\begin{figure}
\begin{center}
\includegraphics[width=1.0\linewidth]{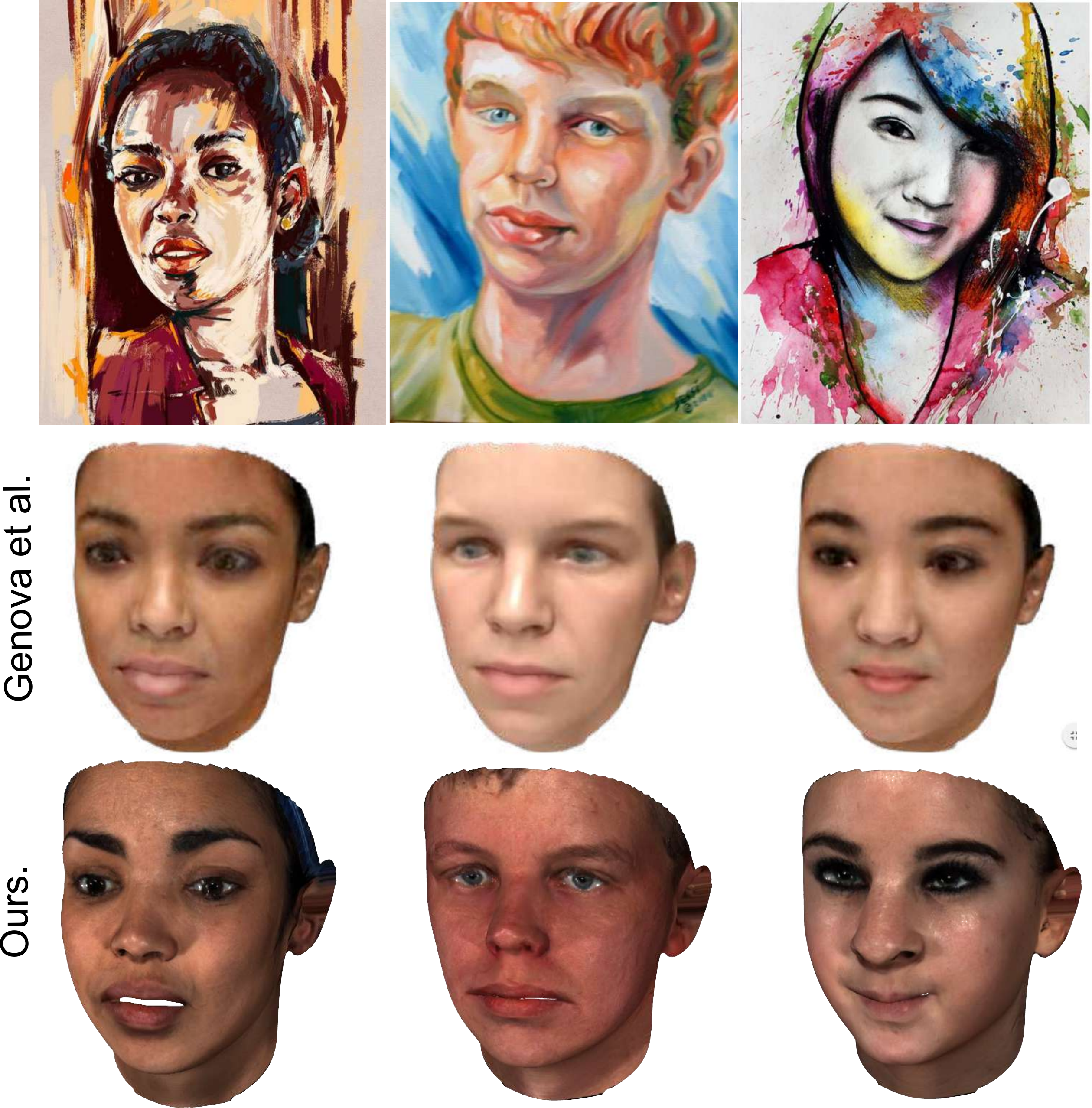}
\end{center}
  \caption{Our results on BAM dataset\cite{wilber2017bam} compared to \cite{genova2018unsupervised}. Our method is robust to many image deformations and even capable of recovering identities from paintings thanks to strong identity features.}
\label{fig:art}
\end{figure}

\section{More Qualitative Results}

\begin{figure}
\begin{center}
\includegraphics[width=1.0\linewidth]{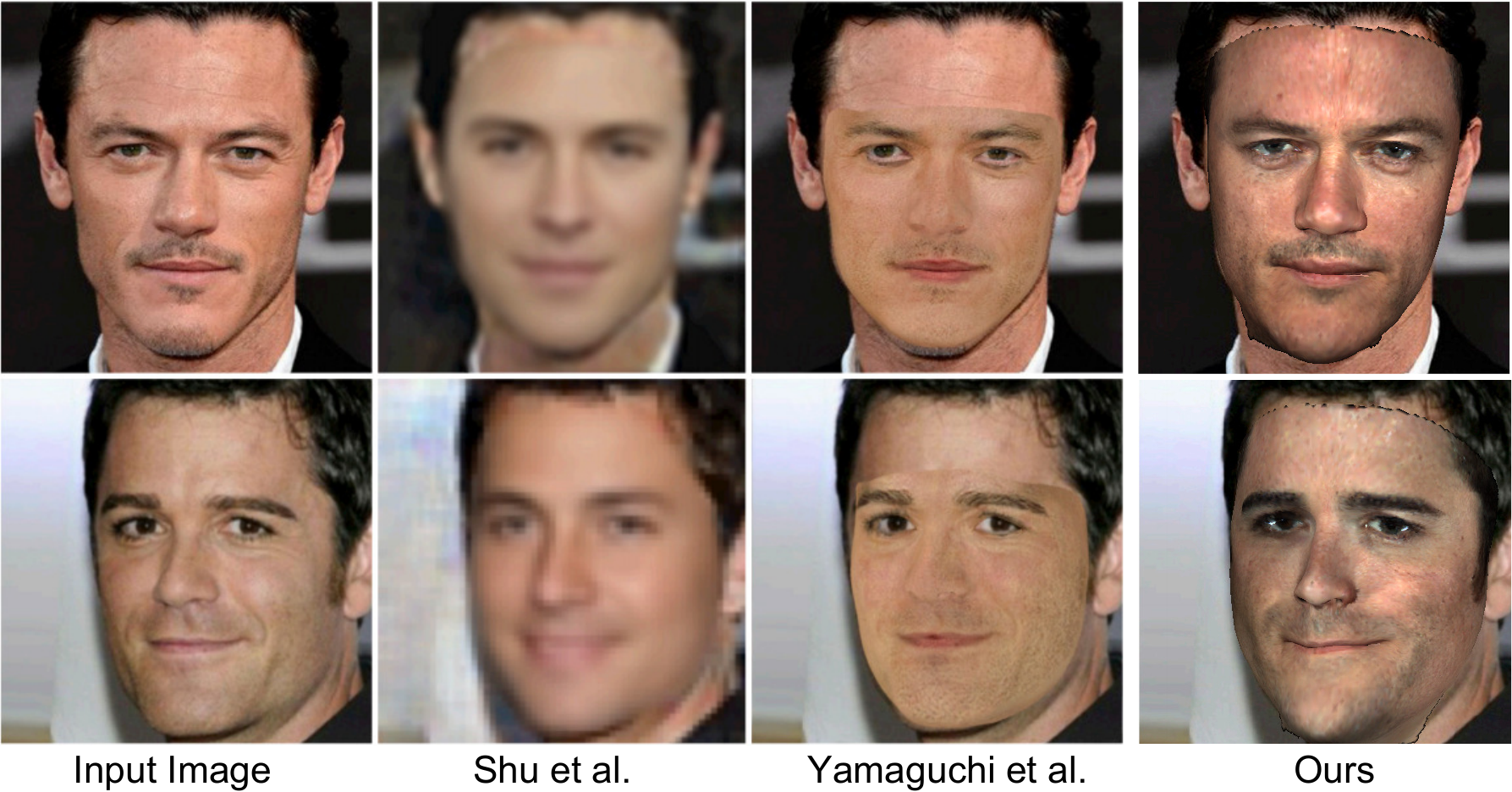}
\end{center}
  \caption{Qualitative comparison with \cite{yamaguchi2018high,shu2017neural} by overlaying the reconstructions on the input images. Our method can generate high fidelity texture with accurate shape, camera and illumination fitting.}
\label{fig:fig1}
\end{figure}

\begin{figure*}
\begin{center}
\includegraphics[width=1.0\linewidth]{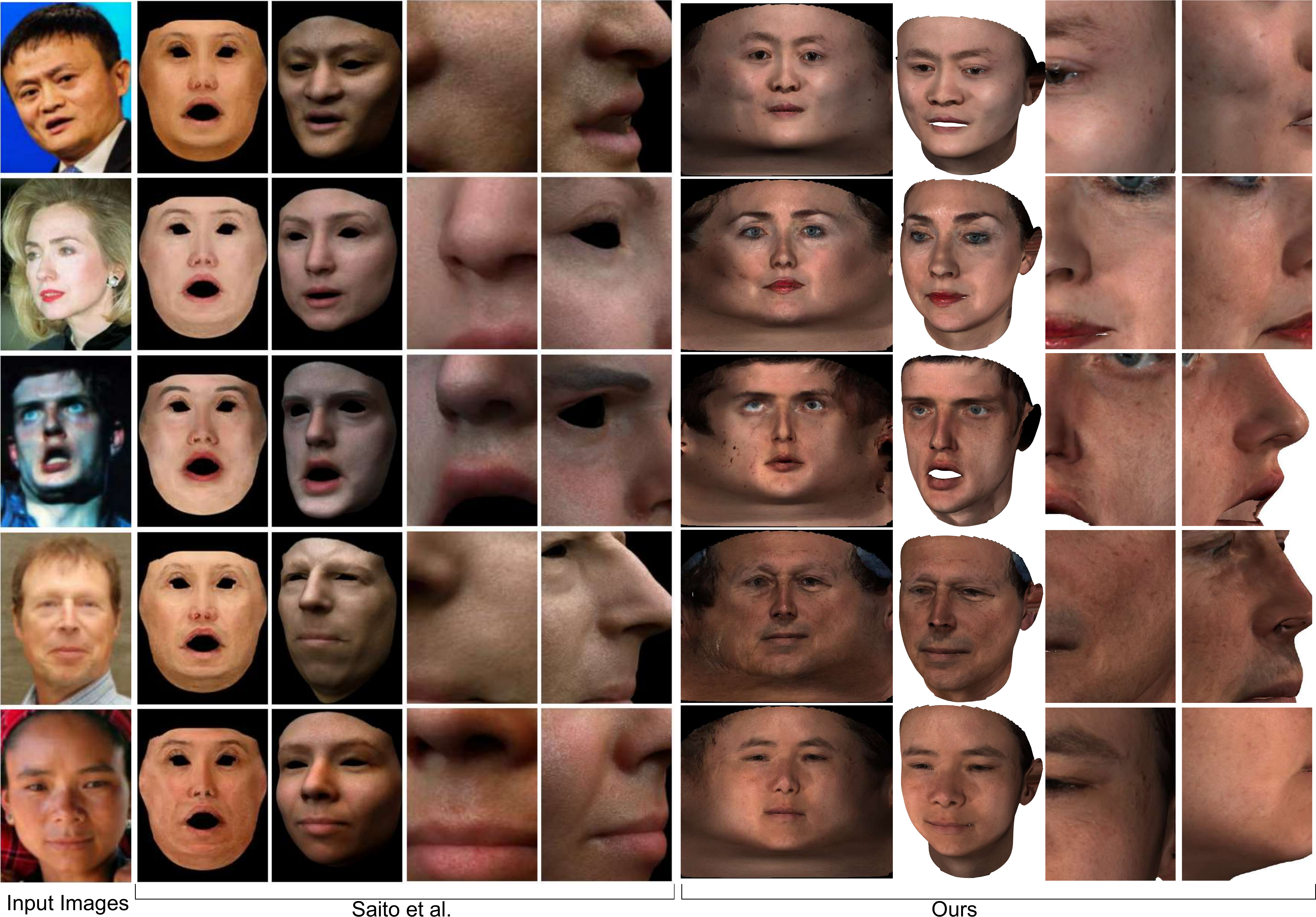}
\end{center}
  \caption{Qualitative comparison with \cite{saito2017photorealistic} by means of texture maps, whole and partial face renderings. Please note that while our method does not require any particular renderer for special effects, e.g., lighting, \cite{saito2017photorealistic} produce these renderings with a commercial renderer called Arnold. }
\label{fig:fig2}
\end{figure*}

Figures \ref{fig:art}, \ref{fig:fig1}, \ref{fig:fig2}, and \ref{fig:fig3}  illustrate the reconstructions of our method under different settings in comparison to the other state-of-the-art methods. Please see figure captions for detailed explanation.


\def \obamavar {0.19}
\begin{figure*}[t]
 \centering 
 \begin{subfigure}{\obamavar\textwidth}
  \includegraphics[width=1.0\textwidth,trim={0 1cm 0 0.5cm},clip]{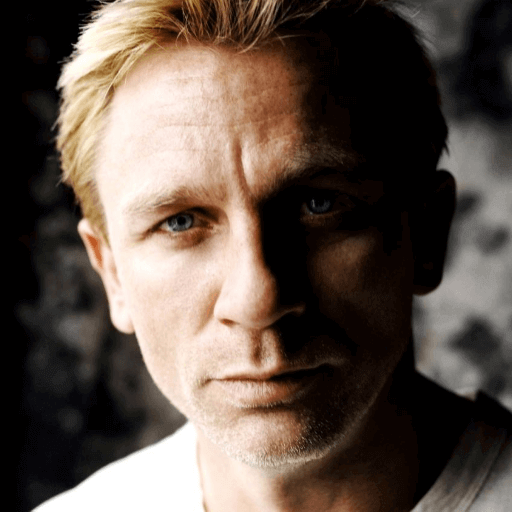}
\end{subfigure}
\begin{subfigure}{\obamavar\textwidth}
  \includegraphics[width=1.0\textwidth,trim={0 1cm 0 0.5cm},clip]{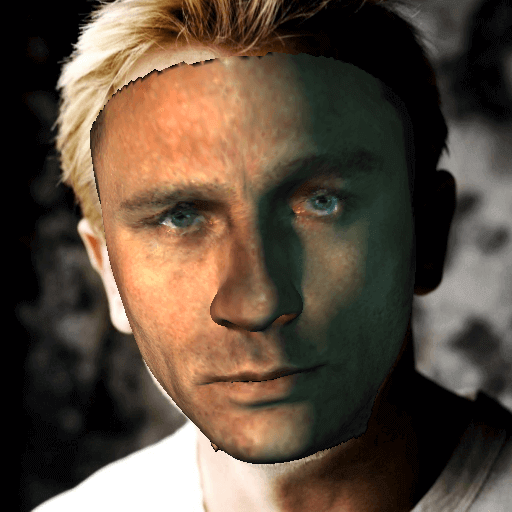}
\end{subfigure}
\begin{subfigure}{\obamavar\textwidth}
  \includegraphics[width=1.0\textwidth,trim={0 1cm 0 0.5cm},clip]{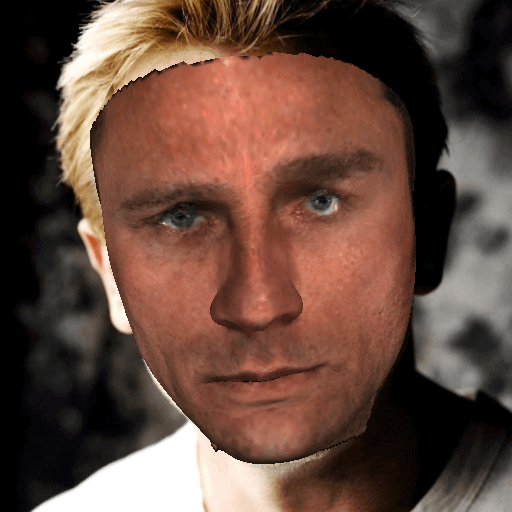}
\end{subfigure}
\begin{subfigure}{\obamavar\textwidth}
  \includegraphics[width=1.0\textwidth,trim={0 1cm 0 0.5cm},clip]{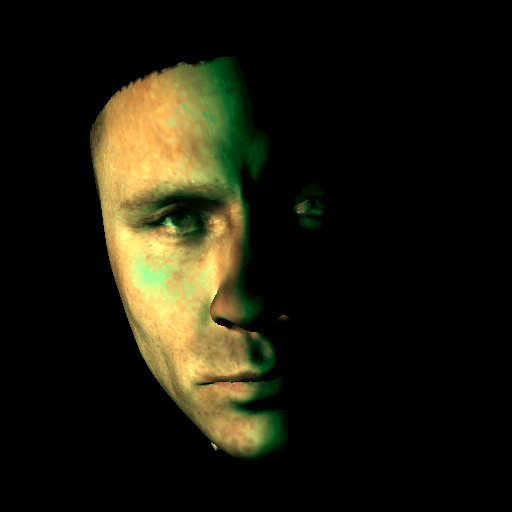}
\end{subfigure}
\begin{subfigure}{\obamavar\textwidth}
  \includegraphics[width=1.0\textwidth,trim={10cm 8cm 5cm 8cm},clip]{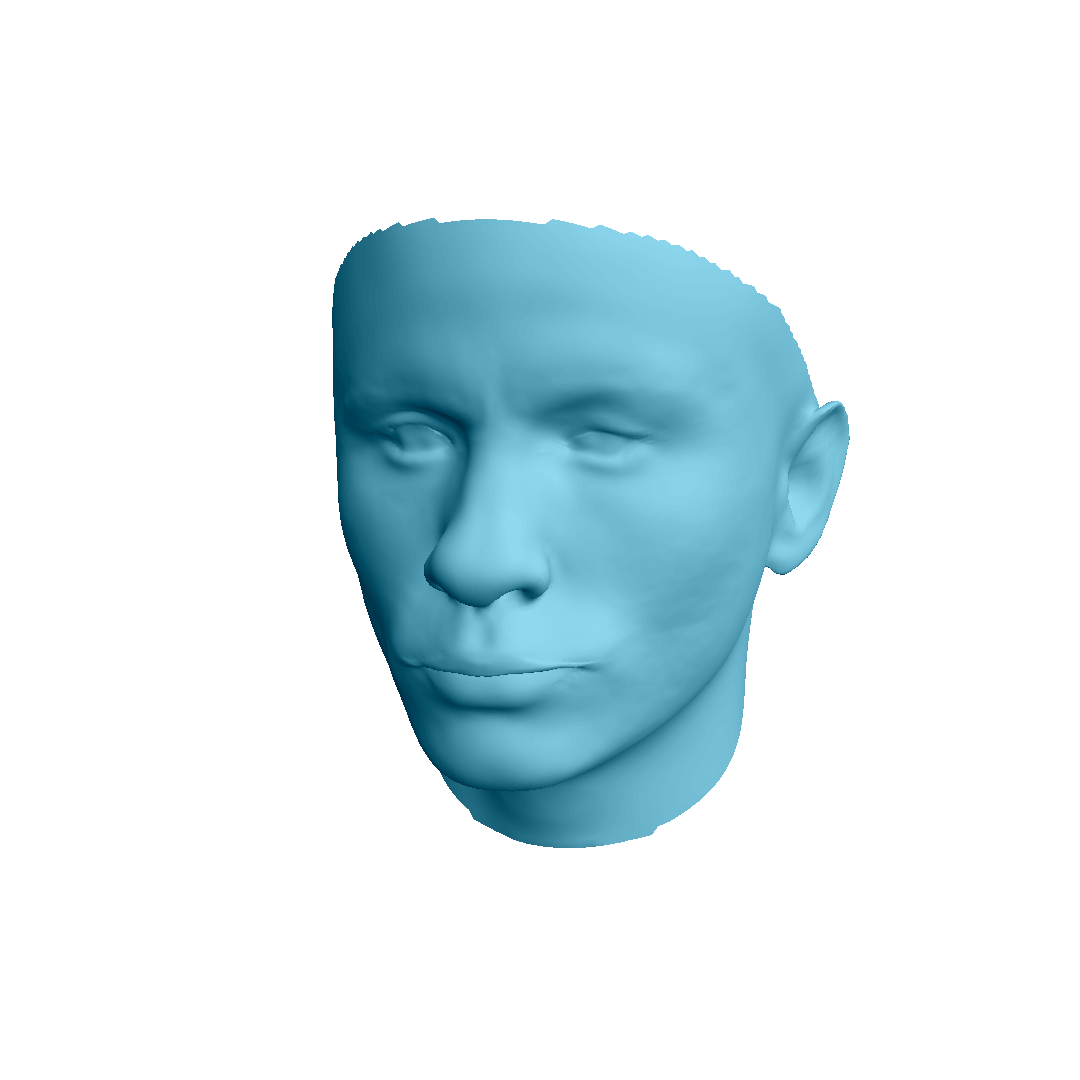}
\end{subfigure}

\begin{subfigure}{\obamavar\textwidth}
  \includegraphics[width=1.0\textwidth,trim={0 1cm 0 0.5cm},clip]{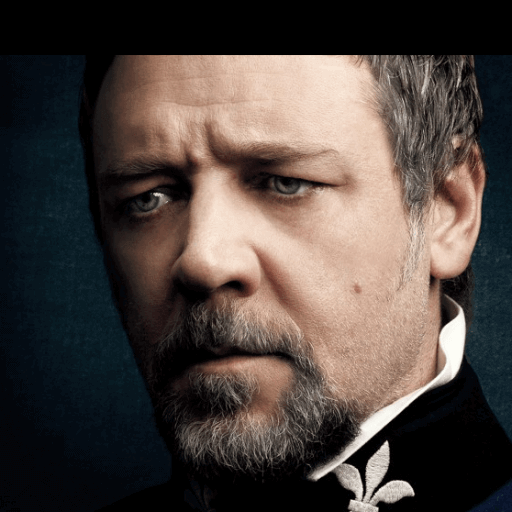}
\end{subfigure}
\begin{subfigure}{\obamavar\textwidth}
  \includegraphics[width=1.0\textwidth,trim={0 1cm 0 0.5cm},clip]{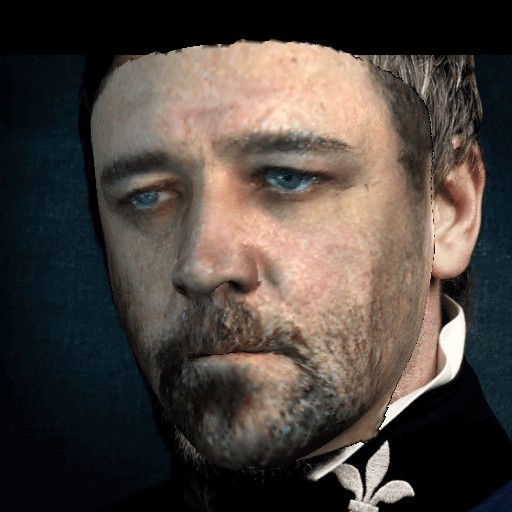}
\end{subfigure}
\begin{subfigure}{\obamavar\textwidth}
  \includegraphics[width=1.0\textwidth,trim={0 1cm 0 0.5cm},clip]{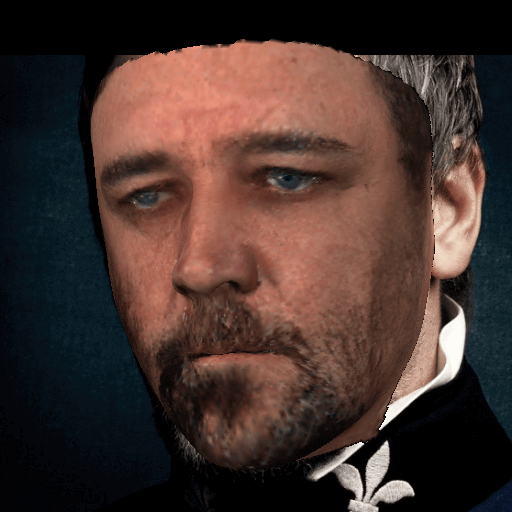}
\end{subfigure}
\begin{subfigure}{\obamavar\textwidth}
  \includegraphics[width=1.0\textwidth,trim={0 1cm 0 0.5cm},clip]{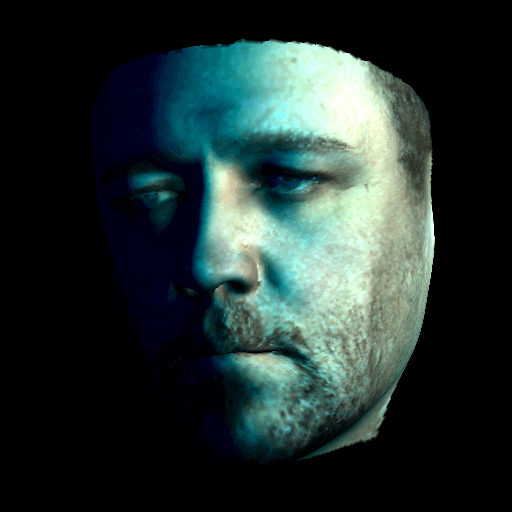}
\end{subfigure}
\begin{subfigure}{\obamavar\textwidth}
  \includegraphics[width=1.0\textwidth,trim={10cm 8cm 5cm 8cm},clip]{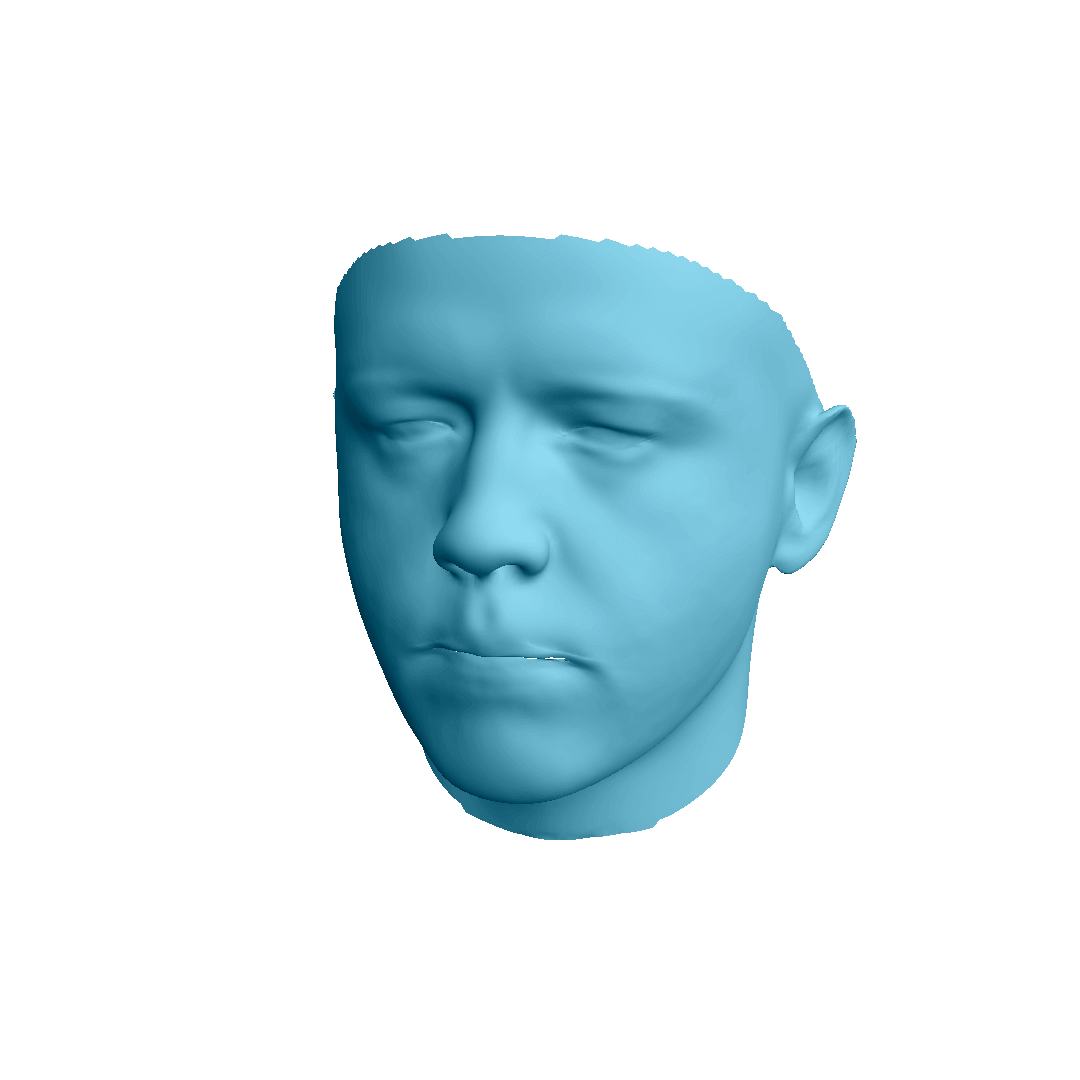}
\end{subfigure}

\begin{subfigure}{\obamavar\textwidth}
  \includegraphics[width=1.0\textwidth,trim={0 1cm 0 0.5cm},clip]{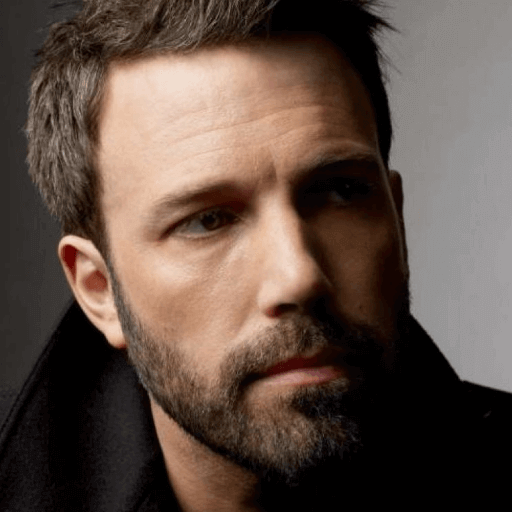}
\end{subfigure}
\begin{subfigure}{\obamavar\textwidth}
  \includegraphics[width=1.0\textwidth,trim={0 1cm 0 0.5cm},clip]{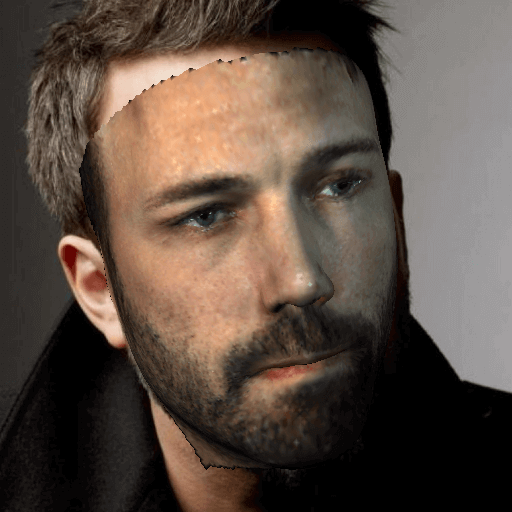}
\end{subfigure}
\begin{subfigure}{\obamavar\textwidth}
  \includegraphics[width=1.0\textwidth,trim={0 1cm 0 0.5cm},clip]{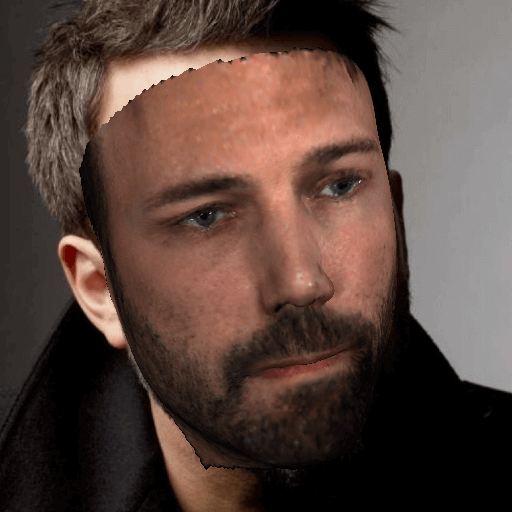}
\end{subfigure}
\begin{subfigure}{\obamavar\textwidth}
  \includegraphics[width=1.0\textwidth,trim={0 1cm 0 0.5cm},clip]{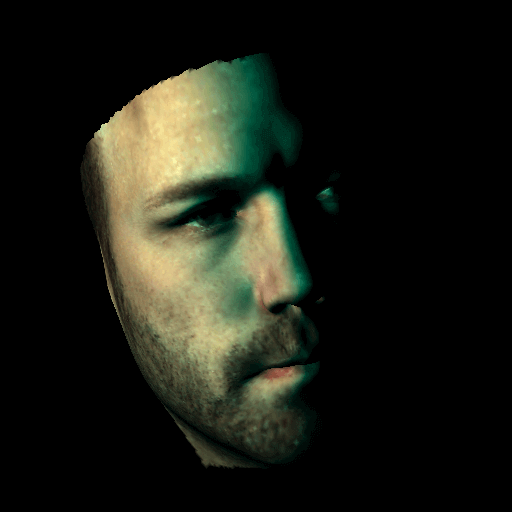}
\end{subfigure}
\begin{subfigure}{\obamavar\textwidth}
  \includegraphics[width=1.0\textwidth,trim={10cm 8cm 5cm 8cm},clip]{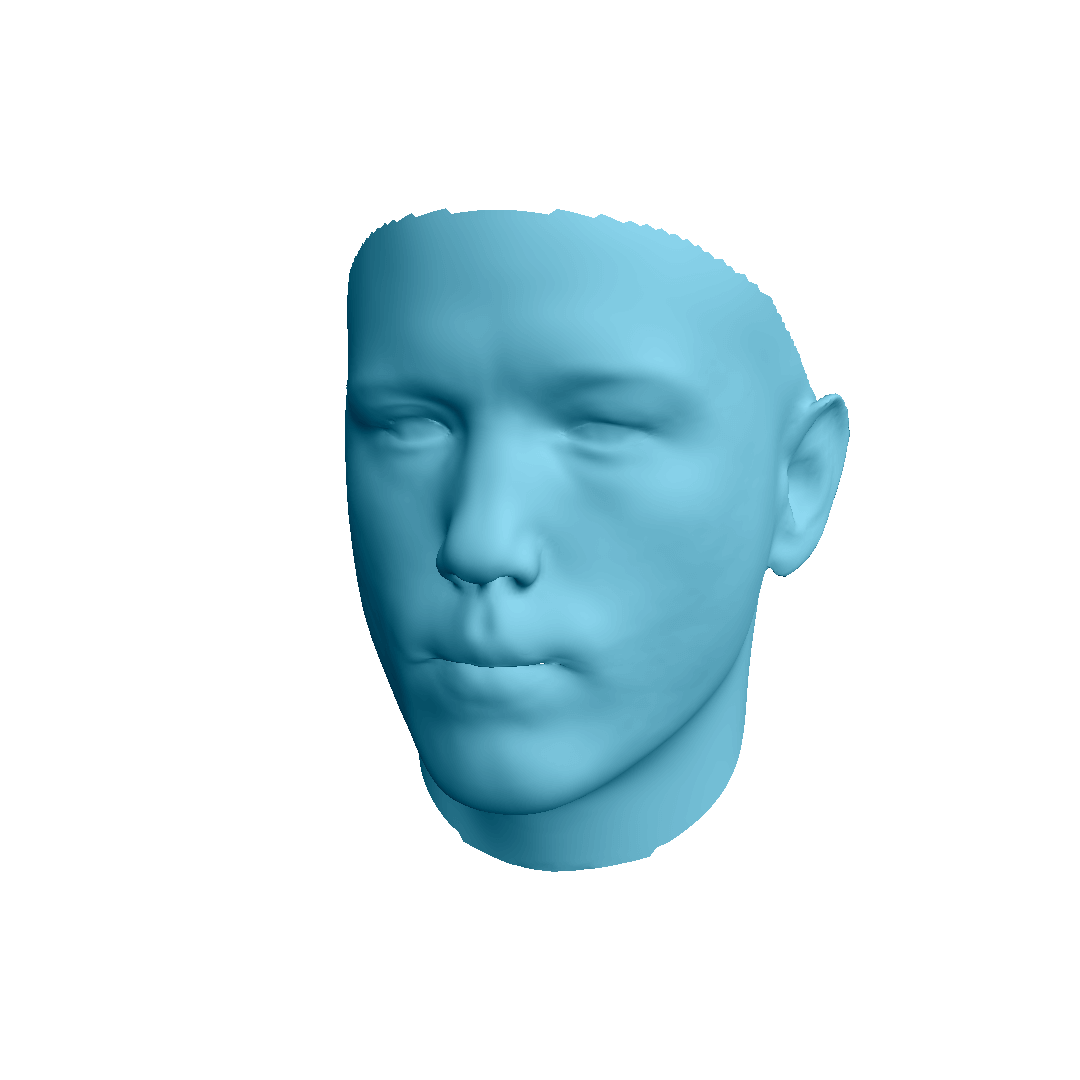}
\end{subfigure}

\begin{subfigure}{\obamavar\textwidth}
  \includegraphics[width=1.0\textwidth,trim={0 1cm 0 0.5cm},clip]{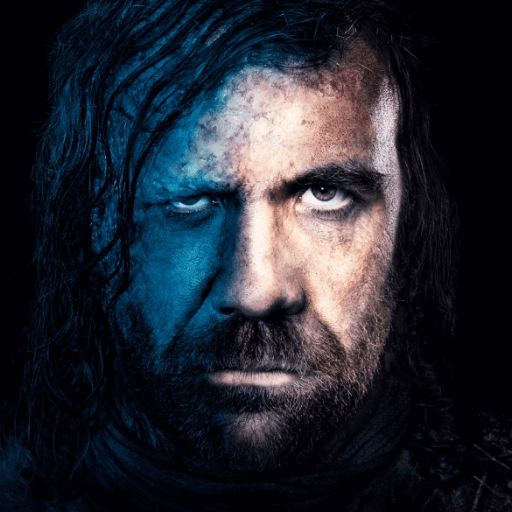}
\end{subfigure}
\begin{subfigure}{\obamavar\textwidth}
  \includegraphics[width=1.0\textwidth,trim={0 1cm 0 0.5cm},clip]{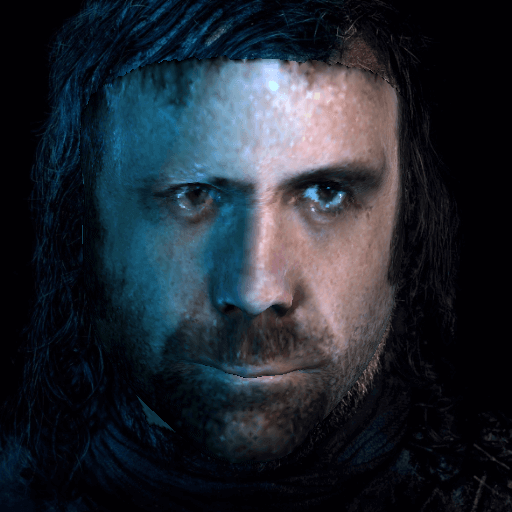}
\end{subfigure}
\begin{subfigure}{\obamavar\textwidth}
  \includegraphics[width=1.0\textwidth,trim={0 1cm 0 0.5cm},clip]{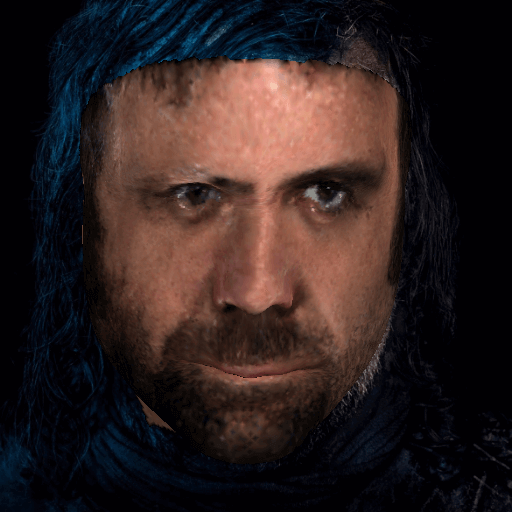}
\end{subfigure}
\begin{subfigure}{\obamavar\textwidth}
  \includegraphics[width=1.0\textwidth,trim={0 1cm 0 0.5cm},clip]{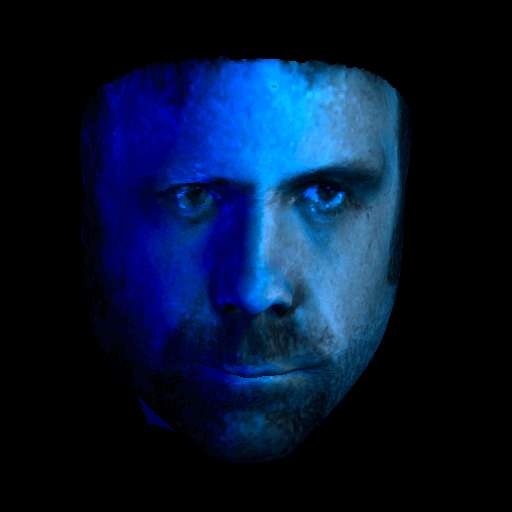}
\end{subfigure}
\begin{subfigure}{\obamavar\textwidth}
  \includegraphics[width=1.0\textwidth,trim={10cm 8cm 5cm 8cm},clip]{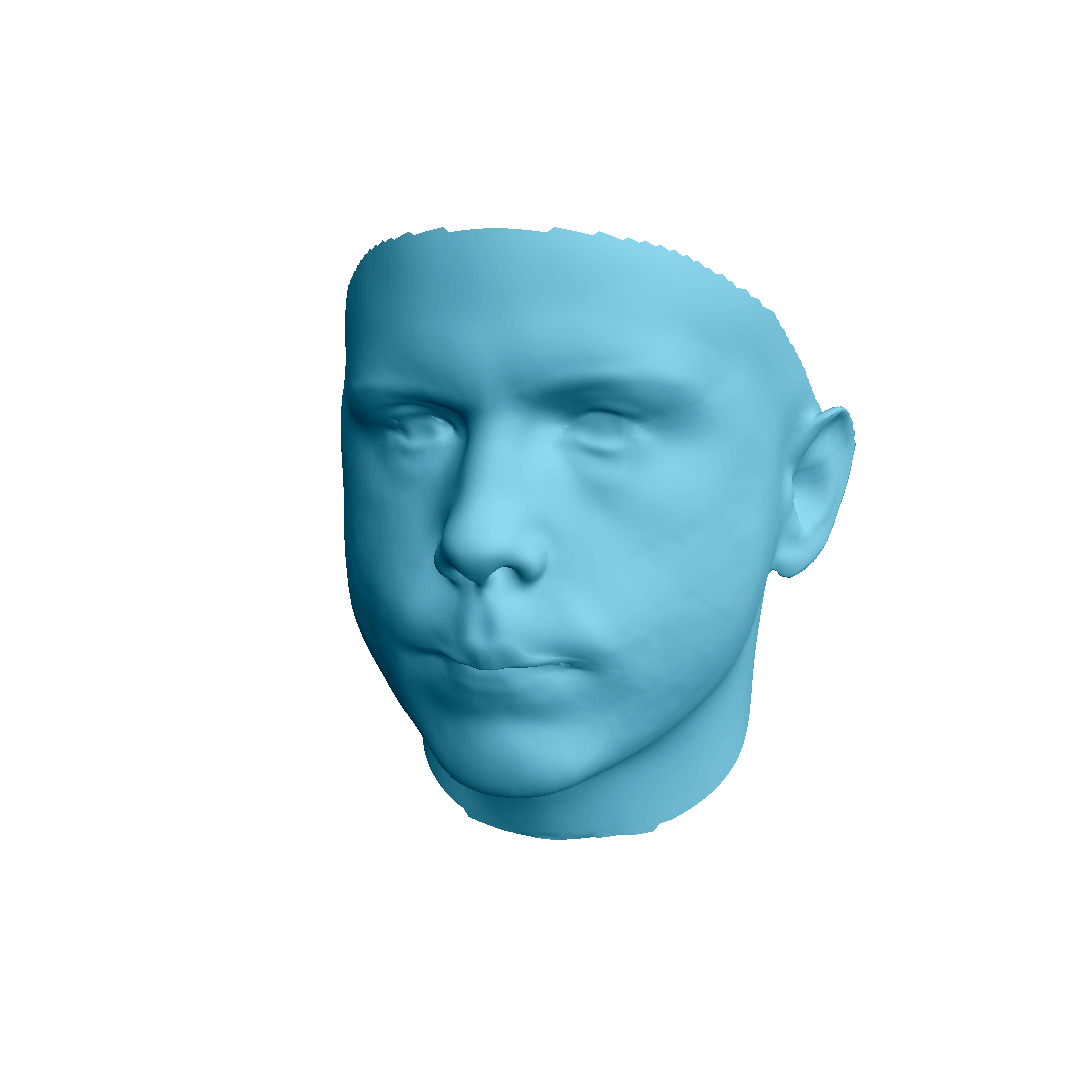}
\end{subfigure}

\begin{subfigure}{\obamavar\textwidth}
  \includegraphics[width=1.0\textwidth,trim={0 1cm 0 0.5cm},clip]{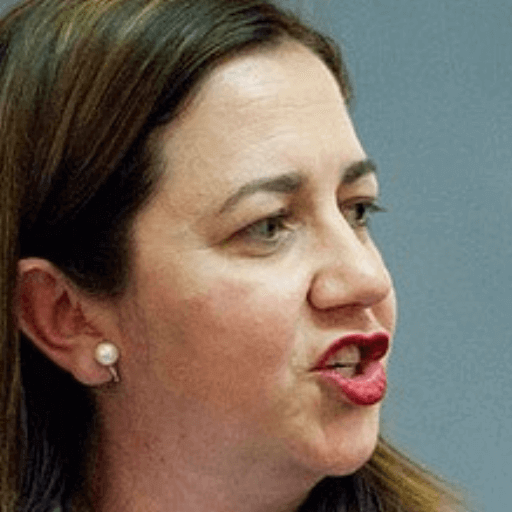}
\end{subfigure}
\begin{subfigure}{\obamavar\textwidth}
  \includegraphics[width=1.0\textwidth,trim={0 1cm 0 0.5cm},clip]{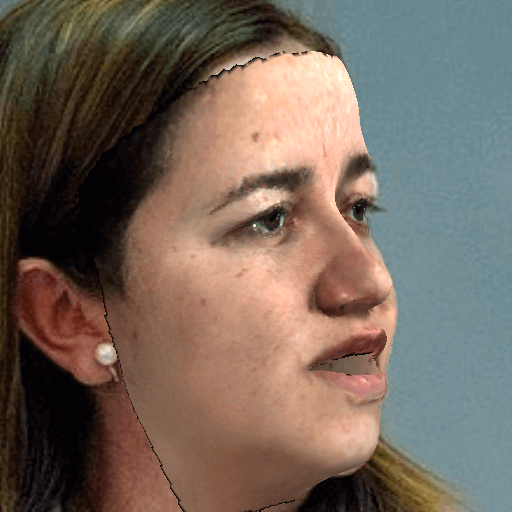}
\end{subfigure}
\begin{subfigure}{\obamavar\textwidth}
  \includegraphics[width=1.0\textwidth,trim={0 1cm 0 0.5cm},clip]{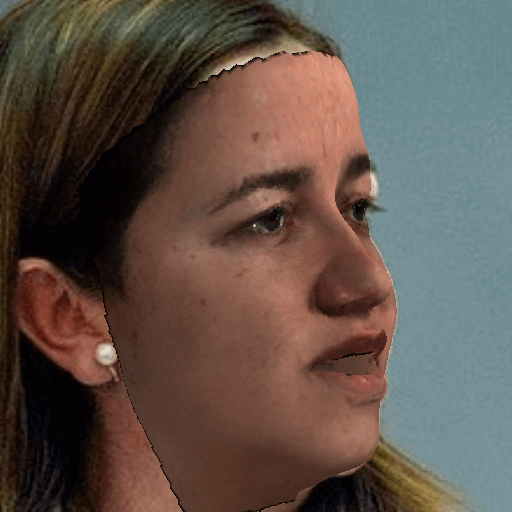}
\end{subfigure}
\begin{subfigure}{\obamavar\textwidth}
  \includegraphics[width=1.0\textwidth,trim={0 1cm 0 0.5cm},clip]{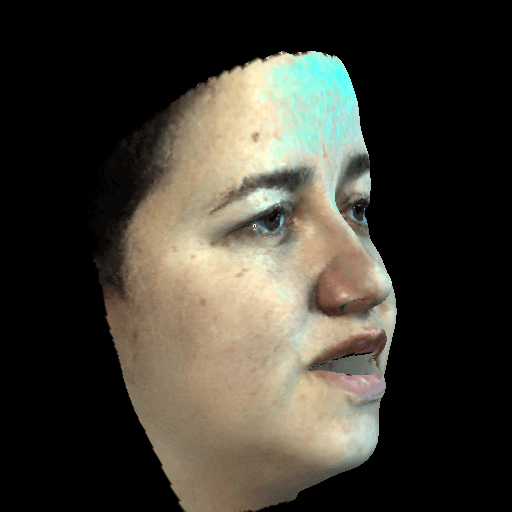}
\end{subfigure}\begin{subfigure}{\obamavar\textwidth}
  \includegraphics[width=1.0\textwidth,trim={10cm 8cm 5cm 8cm},clip]{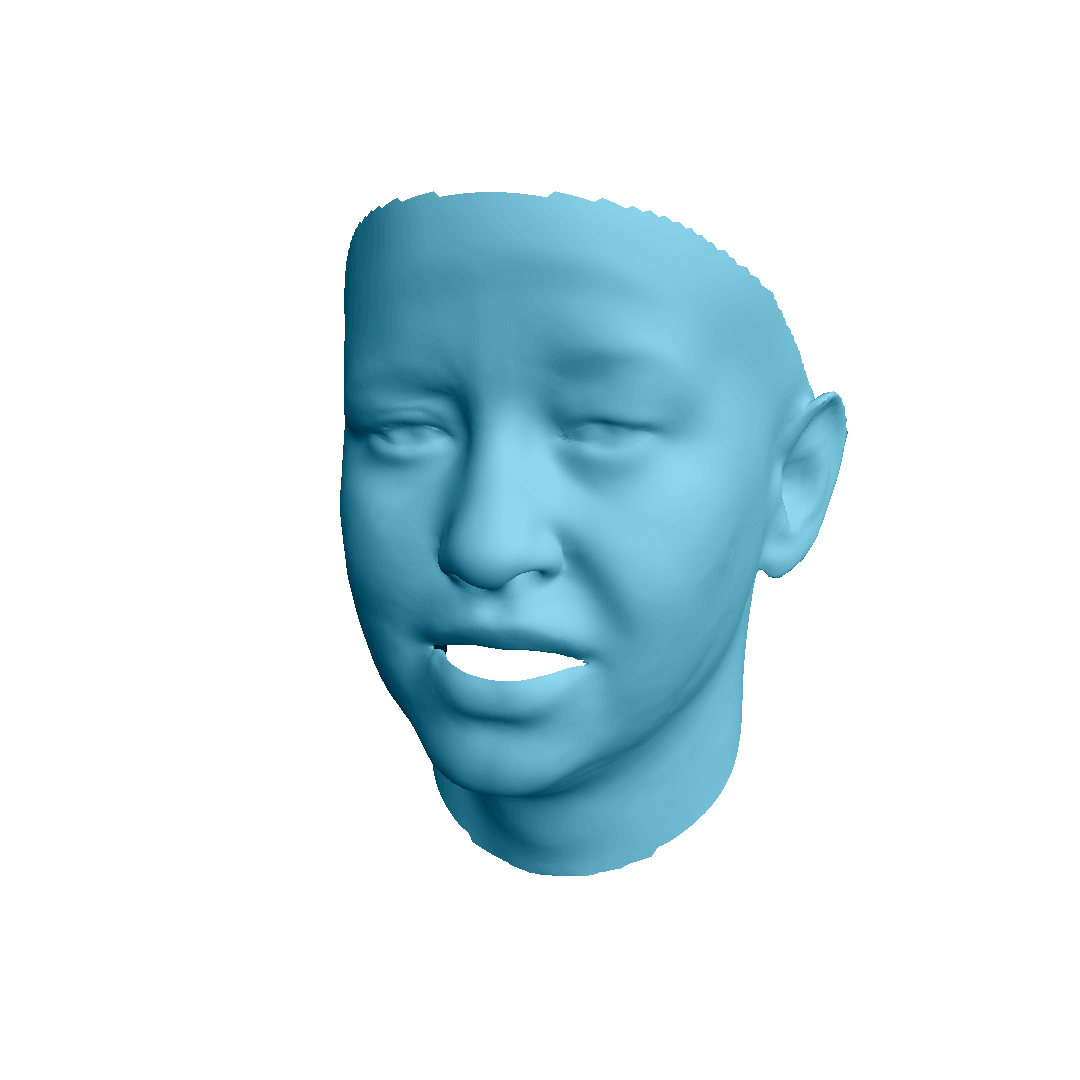}
\end{subfigure}

\begin{subfigure}{\obamavar\textwidth}
  \includegraphics[width=1.0\textwidth,trim={0 1cm 0 0.5cm},clip]{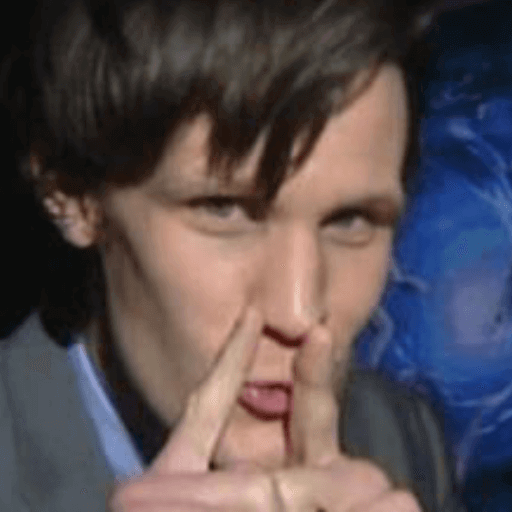}
\tiny\caption{$\mathbf{I^0}$}\end{subfigure}
\begin{subfigure}{\obamavar\textwidth}
  \includegraphics[width=1.0\textwidth,trim={0 1cm 0 0.5cm},clip]{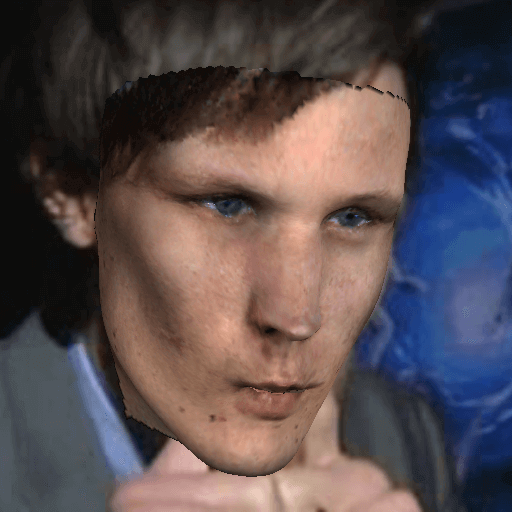}
\tiny\caption{$\mathbf{I^\mathcal{R}}$}\end{subfigure}
\begin{subfigure}{\obamavar\textwidth}
  \includegraphics[width=1.0\textwidth,trim={0 1cm 0 0.5cm},clip]{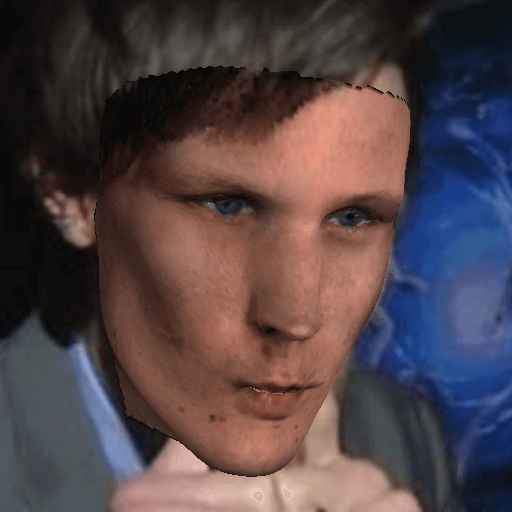}
\tiny\caption{$\mathbf{I^\mathcal{R}_{alb.}}$}\end{subfigure}
\begin{subfigure}{\obamavar\textwidth}
  \includegraphics[width=1.0\textwidth,trim={0 1cm 0 0.5cm},clip]{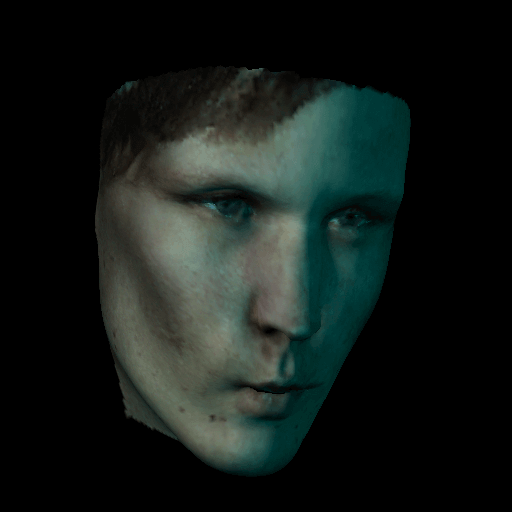}
\tiny\caption{$\mathbf{I^\mathcal{R}}\!\!\!-\! \mathbf{I^\mathcal{R}_{alb.}}$}\end{subfigure}
\begin{subfigure}{\obamavar\textwidth}
  \includegraphics[width=1.0\textwidth,trim={10cm 8cm 5cm 8cm},clip]{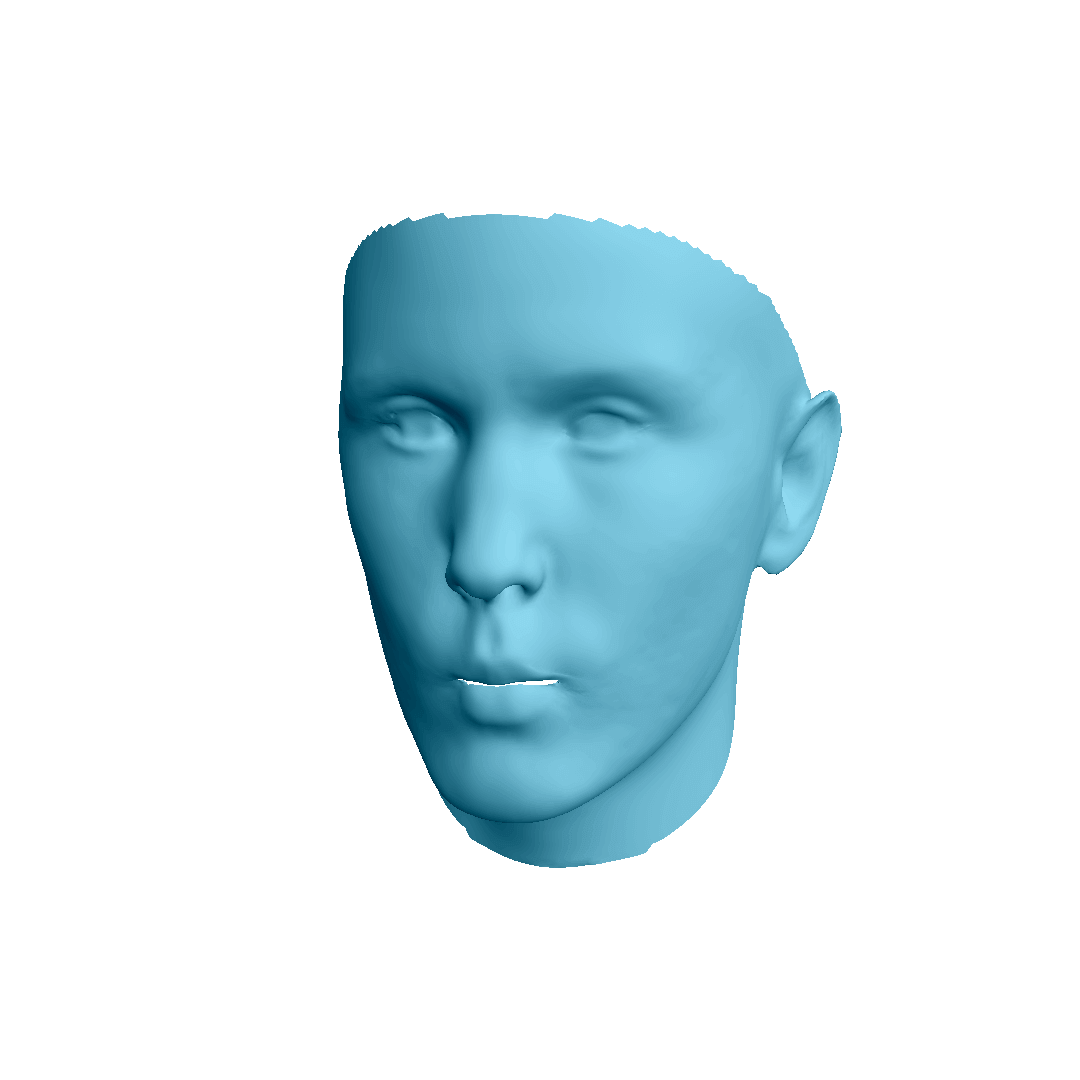}
\tiny\caption{$\mathbf{S}$}\end{subfigure}

\caption{Results under more challenging conditions, \ie strong illuminations, self-occlusions and facial hair. (a) Input image. (b) Estimated fitting overlayyed including illumination estimation. (c) Overlayyed fitting without illumination. (d) Pixel-wise intensity difference of (b) to (c). (e) Estimated shape mesh}
\label{fig:fig3}
\end{figure*}

\end{appendices}
\end{document}